\documentclass{article} 
\usepackage[final]{colm2025_conference}

\usepackage{microtype}
\usepackage{hyperref}
\usepackage{url}
\usepackage{booktabs}

\usepackage{lineno}

\usepackage{graphicx}
\usepackage{url}
\usepackage{ulem}
\usepackage{enumitem}
\usepackage{multirow}
\usepackage{soul}
\usepackage{xcolor}
\usepackage{subcaption} 
\usepackage{amsmath}
\usepackage{amssymb}
\usepackage{mathtools}
\usepackage{amsthm}
\usepackage{caption}
\usepackage[utf8]{inputenc} 
\usepackage[T1]{fontenc}    
\usepackage{amsfonts}       
\usepackage{nicefrac}       

 \usepackage{float}
\usepackage{pifont}
   \definecolor{darkgreen}{rgb}{0.0, 0.5, 0.0}
   \definecolor{darkred}{rgb}{0.5,0.0,0.0}
\newcommand{\greencheck}{{\color{darkgreen}\ding{51}}}

\usepackage{tabularx}
\usepackage{array}

\newcolumntype{C}[1]{>{\centering\arraybackslash}m{#1}}

\usepackage{multibib}

\title{EvidenceBench: A Benchmark for Extracting Evidence from Biomedical Papers}

\author{Jianyou Wang$^1$\thanks{equal contributions} , Weili Cao$^1$\footnotemark[1] , \textbf{Kaicheng Wang$^1$, Xiaoyue Wang$^1$, Ashish Dalvi$^1$, Gino Prasad$^1$, }\\\textbf{Qishan Liang$^3$, Hsuan-lin Her$^3$, Ming Wang$^4$, Qin Yang$^5$, Gene W. Yeo$^3$, David E. Neal$^2$, }\\\textbf{Maxim Khan$^2$, Christopher D. Rosin$^2$, Ramamohan Paturi$^1$\thanks{equal senior authors}, Leon Bergen$^1$}\footnotemark[2] \\
$^1$Laboratory for Emerging Intelligence, University of California, San Diego \\
$^2$Elsevier \\
$^3$Department of Cellular and Molecular Medicine, University of California, San Diego \\
$^4$Sichuan Cancer Hospital \& Institute \\
$^5$The Third People's Hospital of Chengdu \\
\texttt{\{jiw101, w2cao, rpaturi, lbergen\}@ucsd.edu}}

\begin{document}

\ifcolmsubmission
\linenumbers
\fi

\maketitle

\begin{abstract}

We study the task of automatically finding evidence relevant to hypotheses in biomedical papers. Finding relevant evidence is an important step when researchers investigate scientific hypotheses. We introduce EvidenceBench to measure models performance on this task, which is created by a novel pipeline that consists of hypothesis generation and sentence-by-sentence annotation of biomedical papers for relevant evidence, completely guided by and faithfully following existing human experts judgment. We demonstrate the pipeline’s validity and accuracy with multiple sets of human-expert annotations. We evaluated a diverse set of language models and retrieval systems on the benchmark and found that model performances still fall significantly short of the expert level on this task. To show the scalability of our proposed pipeline, we create a larger EvidenceBench-100k with 107,461 fully annotated papers with hypotheses to facilitate model training and development. Both datasets are available at \href{https://github.com/EvidenceBench/EvidenceBench}{https://github.com/EvidenceBench/EvidenceBench}.
\end{abstract}

\section{Introduction}
\label{sec:introduction}

There are more than 1 million biomedical papers currently published per year, and more than 35 million papers collected in the PubMed database of biomedical literature \cite{GonzalezMarquez2024}. The scale of the literature has made it increasingly labor-intensive to determine what is known about a research question. Systems such as OpenAI Deep Research and Elicit aim to automate large-scale analysis of the scientific literature. Notably, it is not enough for these systems to merely produce high-level summaries or restate an article’s claims. To accurately assess what a paper contributes to a research question, such systems should identify the supporting \textit{evidence} -- the experimental or observational data that underpin the claims.

We focus on this critical step of \textbf{evidence retrieval}: given a hypothesis, a system must locate the key parts of a paper that provide pertinent experimental details, numerical findings, or other forms of evidence that address that hypothesis. This step provides grounding to the judgments of an automated system, allowing researchers to quickly judge whether a specific claim is supported by empirical evidence. However, creating and evaluating such systems remain non-trivial. On the one hand, few annotated benchmarks exist in the biomedical domain to measure how effectively models can identify and extract evidence. On the other hand, curating such a dataset is time-consuming and expensive, especially if such annotations need to be performed by experts.

 To address these challenges, we present \textbf{EvidenceBench} and its large-scale extension \textbf{EvidenceBench-100k}, a new benchmark for sentence-level evidence retrieval in biomedical research papers. Our proposed datasets are fully open-sourced under the CC-BY license and encompass a comprehensive range of biomedical topics.

We introduce a novel pipeline for creating EvidenceBench and EvidenceBench-100k, powered by Large Language Models (LLMs). The pipeline has two main components: hypothesis generation and an alignment annotator that matches hypotheses with sentences from papers. Informally, we use existing \textbf{evidence summary} from review papers to generate a hypothesis and use the same \textbf{evidence summary} to find sentences that provide evidence for the generated hypothesis. See Section \ref{sec:dataset description} and Figure \ref{fig:generation} \ref{fig:annotation} for details.

Our pipeline is highly scalable, reducing the construction time of EvidenceBench from over 3,000 human hours and \$120,000 in wages to just 3 API hours and \$5,000 in API costs, using state-of-the-art LLMs at the time of data creation, Claude3-Opus for hypothesis generation and GPT4-0125 for alignment annotation. During the construction of EvidenceBench-100k, we used GPT4-o-mini for alignment annotation and kept the construction time and cost under 24 hours and \$5000. For details about human cost estimate, see Section \ref{ssec:alignment of study aspects and sentences}.

Table \ref{table:fine_tuned} demonstrates EvidenceBench-100k is suitable for fine-tuning LLMs and embedding models as we observed significant improvements of fine-tuned models over their pretrained baselines.

We conduct a benchmarking study on a variety of large language models (LLMs) and embedding models, which provide several insights. First, although LLMs still fall short of expert-level performance on this task, indicating that they cannot replace humans, they have the potential to assist them. Second, embedding models consistently underperform compared to large language models due to lack of global context from research papers. Third, we report that current LLMs trained for long document understanding still get "Lost in the Middle" \citep{lost_in_the_middle}. See Appendix \ref{sec:analysis} for details of our analyses.

We highlight some important sections in the paper: 
\begin{itemize}
    \item Section \ref{ssec:definitions} explains two critical concepts: \textbf{Study Aspect} and \textbf{Source of Information}.
    \item Section \ref{ssec:evaluation metics} presents \textbf{Aspect Recall}, the evaluation metric for our experiments.
    \item Figure \ref{fig:generation} illustrates our key methodological contribution: leveraging expert-written \textbf{evidence summaries} from review papers to guide an LLM-based alignment annotation process (Figure \ref{fig:annotation}), ensuring our pipeline adheres to human expert judgments.
    \item Section \ref{ssec:expert validation of hypotheses} shows experts confirm the scientific value of our generated hypotheses.
    \item Section \ref{ssec:alignment of study aspects and sentences} and Table \ref{tab:hypothesis_test} provide statistical evidence that our alignment annotation procedure achieves comparable performance to biomedical PhD students.
\end{itemize}

\section{Task Formulation}
\label{sec:task formulation}

The EvidenceBench task is to identify the most important pieces of evidence relevant to a hypothesis. This is formulated as a sentence retrieval task. Given a paper, the task is to retrieve a set of sentences that jointly provide the most important pieces of evidence. See Figure \ref{fig:evaluation} for an example.

\subsection{Definitions}
\label{ssec:definitions}

\textbf{Candidate Pool:}
The full-text of a research paper is presented as a list of sentences \footnote{\small{Figures and tables are excluded for two reasons. First, EvidenceBench is designed to evaluate textual models. Second, important information from figures and tables is often restated in the paper's text. In the rare case where an evidence summary is largely based on figures or tables from a paper, it means most study aspects cannot be covered by sentences from the paper. Such cases will be discovered at the alignment annotation stage and filtered from EvidenceBench. Specifically, we filter out cases where less than 70\% of study aspects from the evidence summary are covered by sentences.}}.  This ordered list of sentences is the Candidate Pool. Very importantly, research papers and review papers are completely different. 

\textbf{Evidence Summary:}
An evidence summary is written by human experts. It is included in open-sourced review papers, such as surveys, monographs and systematic reviews. An evidence summary is directly linked to one single research paper. It contains all pieces of evidence (from this research paper) that the human experts believe are important and relevant to a hypothesis. Evidence summary could take the form of a normal paper summary, or a bulleted list, or even tabular format.

\textbf{Study Aspect:}
A study aspect, denoted as $f_i$, is a single piece of information. For instance, an experimental outcome or a detail from study design. Note, study aspects are decomposed from a evidence summary written by experts. Each study aspect is identified by human experts and represent an important detail relevant for a hypothesis. Therefore, study aspects are considered to be human experts' judgment. See Figure \ref{fig:evaluation} for example study aspects.

\textbf{Source of Information:}
A sentence in a research paper is considered a source of information for a study aspect if it satisfies the following two criteria. 
\begin{enumerate}
    \item The content of the sentence implies most of the study aspect.
    \item For any part of the study aspect that the sentence does not cover, the information must be easily deducible from the surrounding context.
\end{enumerate}
Given a sentence $s_j$ and a study aspect $f_i$, define the source-of-information indicator function $\mathcal{S}(f_i, s_j)$:
\[
\mathcal{S}(f_i, s_j) = 
\begin{cases} 
1 & \text{if } s_j\text{ is a the source of information for } f_i \\
0 & \text{otherwise}
\end{cases}
\]
Note, if a sentence is a source of information for a study aspect, we informally say the sentence covers this aspect. Note, since study aspect is decomposed and only represent one piece of information, one single sentence is enough to cover it.

\textbf{Hypothesis:}
A hypothesis is a scientific generalization, usually expressible in one line (less than 50 words). It should not be tied to specific details of an experiment. See Figure \ref{fig:evaluation} for an example and Section \ref{ssec:hypothesis collection} for the generation and validation process.

\textbf{Evidence Set:}
The evidence set is the set of study aspects which provides evidence relevant to a hypothesis, according to human expert judgments. Very importantly, we use an \textbf{Evidence Summary} to derive an evidence set, see Figure \ref{fig:generation} right side. A study aspect in this set may provide evidence on its own, or only in combination with other study aspects.

\subsection{Task Definition}
\label{ssec:task definition}

We now introduce our primary task, \textbf{Evidence Retrieval @K (ER@K)}. Informally, the task is to find K sentences in a research paper which provide the greatest amount of evidence relevant to a hypothesis. This is operationalized as finding sentences in the research paper which cover the most study aspects from the evidence set.

Formally, given a hypothesis and a candidate pool, the task for a system is to retrieve K sentences from the candidate pool which provide evidence relevant to the hypothesis. The retrieved sentences are then evaluated against the evidence set, which contains ground-truth study aspects (i.e. pieces of evidence relevant to the hypothesis  identified by human experts). The goal is for the K retrieved sentences to be sources of information for as many of the study aspects in the evidence set as possible. During the retrieval task, the system does not have access to the ground-truth evidence set; the evidence set is only used for evaluation.

In the second version of the task, only study aspects related to the results and analyses are considered. Study aspects related to background and methods are filtered out of the evidence set. This task is called \textbf{Result-ER@K}, and focus on system's ability to identify numerical and experimental results.

\subsection{Evaluation Metrics}
\label{ssec:evaluation metics}

To determine the quality of a system's retrieved sentences, we use
\textbf{Aspect Recall}.
Let $\{s_1, \dots, s_k\}$ be the set of retrieved sentences, and $\{f_1, \dots, f_m\}$ be the set of study aspects in the evidence set. The Aspect Recall is defined as

\begin{equation}
\frac{\sum_{f_j} \displaystyle \mathbf{1}_{\left(\sum_{s_i} \mathcal{S}(s_i,f_j)\right) \geq 1}}{m}
\label{eq:aspect_recall}
\end{equation}

\begin{figure}
    \centering
    \includegraphics[width=\linewidth]{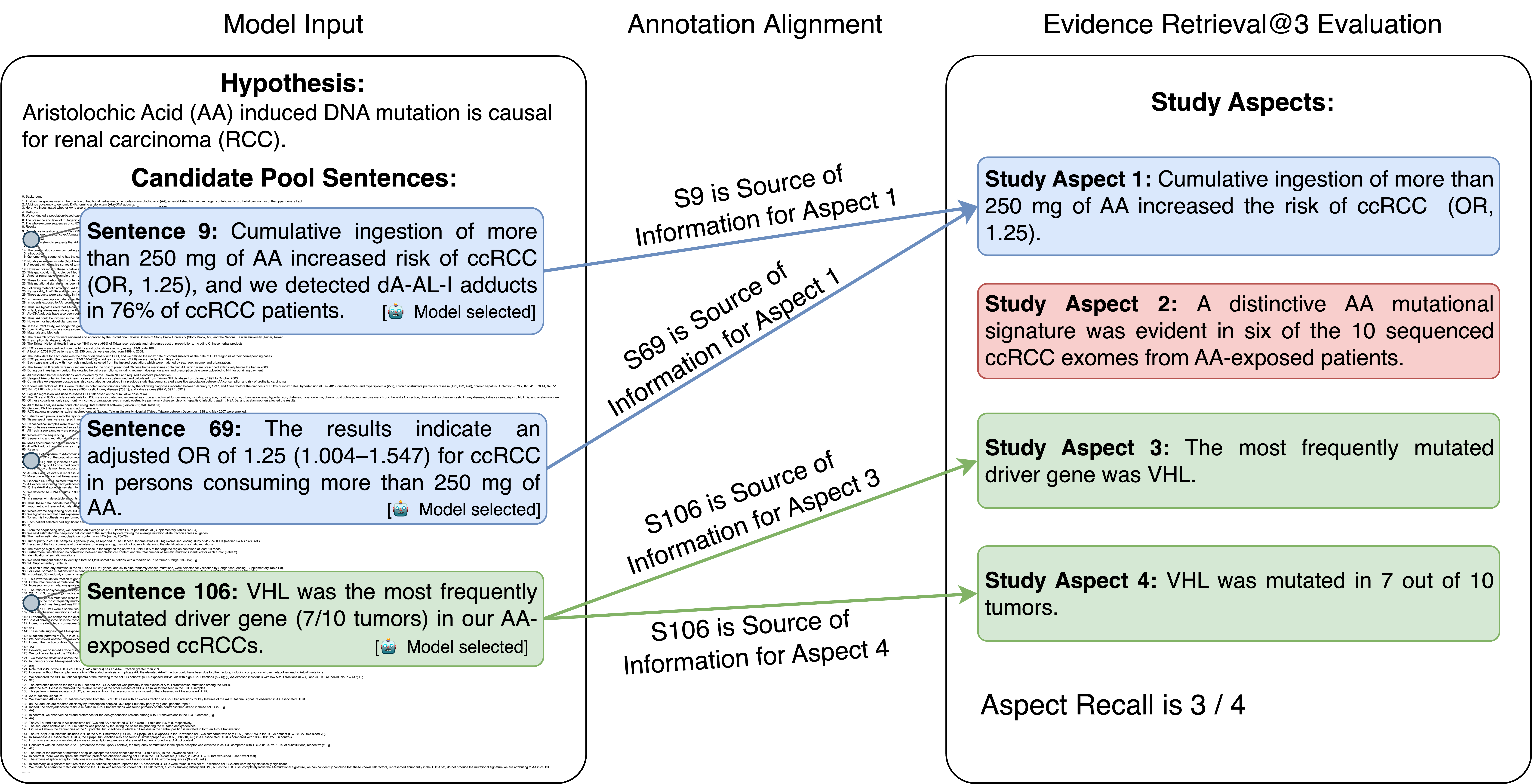}
    \caption{\small{In the task of ER@3, a model sees a hypothesis and the full sequence of sentences from a paper as candidate pool and must select up to 3 sentences. The model selects S9, S69, and S106 as the set of retrieved sentences. When compared against the ground-truth evidence set which contains 4 aspects, these 3 sentences only cover Aspect 1, 3, and 4, since S9 and S69 are repetitive and cover the same Aspect 1. Aspect 2 is missed, resulting in 75\% Aspect Recall.}}
    \label{fig:evaluation}
\end{figure}

This measures the fraction of study aspects that can be covered by a retrieved set of k sentences. See Figure \ref{fig:evaluation} for an example calculation of Aspect Recall \footnote{\small{Our objective is to find the smallest amount of sentences to cover the maximum amount of study aspects. We choose Aspect Recall over precision metrics because precision metrics could inadvertently encourage models to select redundant sentences, which is counterproductive to our objective.}}.

\section{Dataset Construction Pipeline}
\label{sec:dataset description}

\subsection{Data Sources}
\label{ssec:data sources}
There are two data sources for EvidenceBench and EvidenceBench-100k. First, a collection of 107,887 CC-BY open-sourced biomedical research papers where each research paper represents a datapoint. Second, a collection of 44,772 review papers from PubMed Central. Each biomedical research paper has a corresponding evidence summary included in a review paper. Specifically, EvidenceBench has 426 datapoints and EvidenceBench-100k has 107,461 datapoints. Our datasets cover a wide range of biomedical topics, see Appendix \ref{ssec:appendix_dataset coverage analysis} for details.

\subsubsection{Train/Test Split}
\label{ssec:train/test split}

We use a random train/test split of 133/ 293 task instances for the original EvidenceBench. We use a random train/test split of 87,461/20,000 for EvidenceBench-100k. All of the prompt optimization is performed on the train sets and all of the few-shot examples used in the prompts are selected from the train sets. All datasets have CC-BY licenses.

\begin{table}[htbp]
\caption{\small{EvidenceBench Test Set. Optimal Number of Sentences refers to the smallest number of sentences that are sources of information for the most of study aspects in an evidence set.}}
\label{tab:testset statistics}
\centering
\scriptsize
\setlength{\tabcolsep}{4pt}
\begin{tabular}{lc|ccc|ccc|ccc|p{1cm}p{1cm}c}
\toprule
                            & & \multicolumn{3}{c|}{\textbf{Candidate Tokens}} & \multicolumn{3}{c|}{\textbf{Sentences}} & \multicolumn{3}{c|}{\textbf{Study Aspects}} & \multicolumn{3}{c}{\textbf{Optimal Number of Sentences}} \\
                    Dataset & n &    min &     avg &   max &       min &    avg & max &             min &  avg & max &   min &  avg & max \\
\midrule
                   Test Set &  293 &   1691 &  5578 & 23980 &        48 & 168.1 & 794 &               2 &  9.5 &  36 &     1 &  4.6 &  18 \\

Test Set (Result Retrieval) &  288 &   1691 &  5582 & 23980 &        48 & 168.4 & 794 &               1 &  4.2 &  18 &     1 &  2.1 &   7 \\
\bottomrule
\end{tabular}

\end{table}

\subsection{Dataset Pipeline Overview}
\label{ssec:dataset pipeline overview}

A task instance (i.e. a datapoint) is constructed as follows. After we harvest an expert-written evidence summary from a review paper, we generate a hypothesis from it. Since the evidence summary was written to summarize a research paper, we take the sentences from this research paper as the candidate pool for the task instance. We then decompose the expert-written evidence summary into a set of study aspects, also known as the evidence set. Each study aspect is used to annotate each sentence in the candidate pool. This is the alignment annotation process, which determines sentences in the research paper that are sources of information for the study aspect, and consequently, are relevant for the hypothesis. In the next sections, we explain the procedures of harvesting evidence summary from review papers, hypothesis generation, aspect decomposition, and alignment annotation.

\begin{figure}[t]
    \centering
    \includegraphics[width=\linewidth]{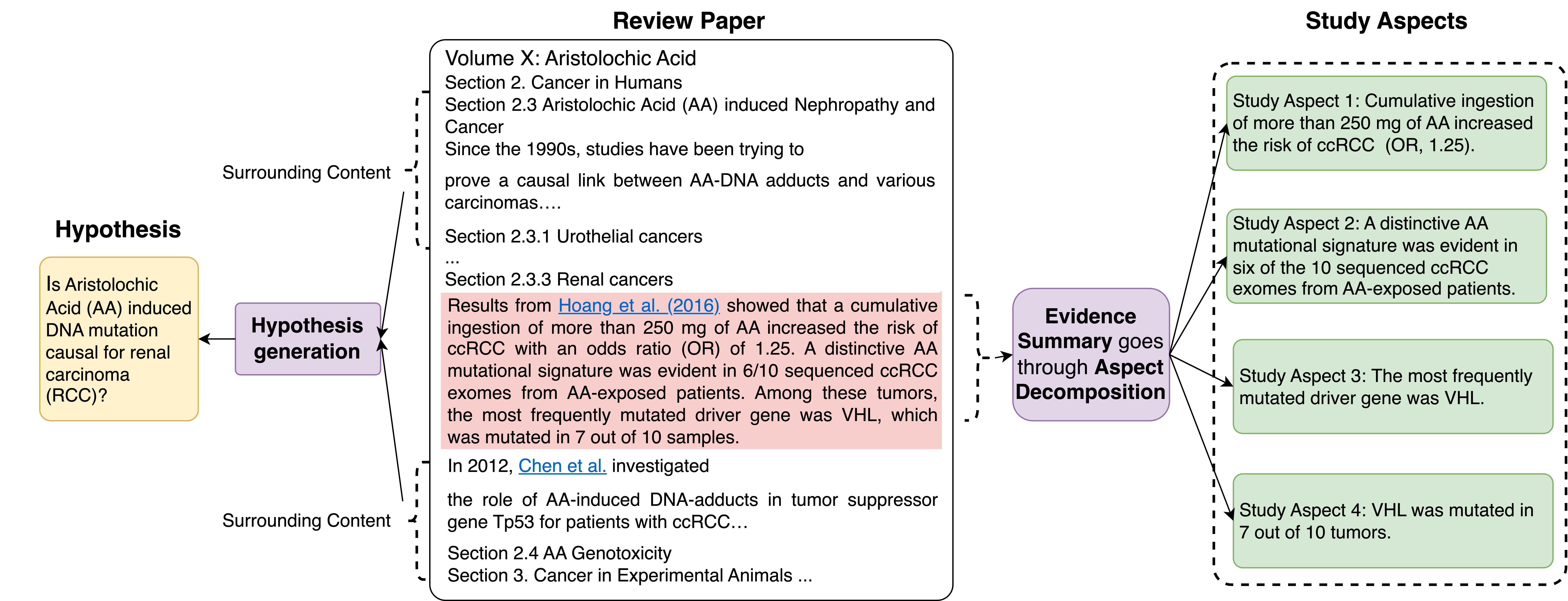}
    \caption{\small{The highlighted paragraph in the review paper is the evidence summary. The left side shows how a hypothesis is generated (extracted) from the evidence summary and its surrounding context in the review paper. The right side shows the evidence summary being decomposed into study aspects.}}
    \label{fig:generation}
\end{figure}

\textbf{Harvesting the Evidence Summary:}
The highlighted portion in the review paper in Figure \ref{fig:generation} is an evidence summary. Evidence summary has an XML citation embedded in it, so it can be identified and harvested by a deterministic algorithm. 
See Appendix \ref{sec:appendix data preprocessing} for details.

\textbf{Aspect Decomposition:}
For each evidence summary, we decompose it into study aspects. These study aspects comprise the evidence set for the research paper. Table \ref{tab:testset statistics} shows a summary on average contains 10 study aspects and half of them are results aspects. In EvidenceBench, decomposition is done by GPT4-0125 and inspected by human researchers. In EvidenceBench-100k, the decomposition is done by GPT-4o-mini-0718 and 200 randomly sampled instances are inspected by human researchers and are found to be of high quality.

\subsection{Hypothesis Generation}
\label{ssec:hypothesis collection}

A review paper focuses on a specific hypothesis and survey a number of research papers, summarizing the evidence that each provides for the hypothesis. For each evidence summary, our goal is to extract the hypothesis that it is providing relevant evidence to. Think of a hypothesis as an explicit or implicit question waiting to be recovered. In order to do this, we provide an LLM (Claude3-Opus) with the evidence summary as well as surrounding paragraphs. The model is then prompted to recover the hypothesis being discussed in the review paper. See Figure \ref{fig:generation} left side. To ensure high quality hypotheses for EvidenceBench-100k, Claude3-Opus is also used.

\subsubsection{Expert Validation of Hypotheses}
\label{ssec:expert validation of hypotheses}
We perform an expert evaluation of the hypotheses extracted from review papers in EvidenceBench, focusing on two questions:
\begin{enumerate}
\item Does the hypothesis have sufficient scientific value?
\item Does the corresponding evidence summary provide evidence which is relevant to the hypothesis?
\end{enumerate}

The annotation team for this task consisted of three medical doctors. The first expert defined the annotation guidelines and provided feedback on an initial set of 20 extracted hypotheses. This feedback was also used to perform prompt optimization for Claude3-Opus.

After finalizing the guidelines and prompt, a separate set of 50 hypotheses was generated. The two other annotators each evaluated 25 hypotheses. Annotation guidelines are in Appendix \ref{sec:appendix_annotation guideline}.

For Question 1, 50/50 hypotheses were judged to have sufficient scientific value. For Question 2, 47/50 hypotheses were judged to be relevant to the corresponding evidence summaries. This demonstrates that hypotheses were correctly extracted from review papers.

\subsection{Alignment Annotation of Study Aspects and Sentences}
\label{ssec:alignment of study aspects and sentences}

We have so far described the procedure for decomposing study aspects and recovering hypotheses from the review papers. A list of study aspects describes the evidence that a specific research paper provides relevant to a hypothesis. The final step is to identify which sentences of the original research paper serve as sources of information for each study aspect. Because study aspects can, in general, come from any part of the research paper, this requires annotating every sentence in the research paper according to whether it matches each study aspect. 

Table \ref{tab:testset statistics} shows that each paper has approximately 168 sentences and 10 study aspects on average. EvidenceBench contains more than 400 research papers. Sentence-by-sentence annotation requires approximately 700,000 sentence annotations, which is infeasible given the use of expert annotators; we estimate that it would require more than 3000 hours of annotation \footnote{\small{Our experiments showed that a single bioinformatics PhD student cannot reliably annotate one paper for one aspect in less than 20 minutes. Reliable annotation requires two Bio PhD students collaborating, taking 20-30 minutes per aspect. Only with this collaborative approach did annotations show high inter-annotator agreement across different teams. The calculation breaks down as follows: 2 PhD students * 25 minutes/aspect * 10 aspects/paper * 400 papers = 3,333 human hours. Each PhD level expert has an expected hourly wage between \$40 to \$95 (see wage standard in \cite{gpqa}).}}. We therefore develop a pipeline for automating the annotation process, and perform human evaluation of its reliability. We observe that the task of labeling a sentence according to whether it is a source of information for a study aspect is considerably simpler than the benchmark's full sentence retrieval task. It only requires a judgment of whether a single sentence from the research paper contains most of the same information as a study aspect. 

\begin{figure}[htbp]
    \centering
    \includegraphics[width=\linewidth]{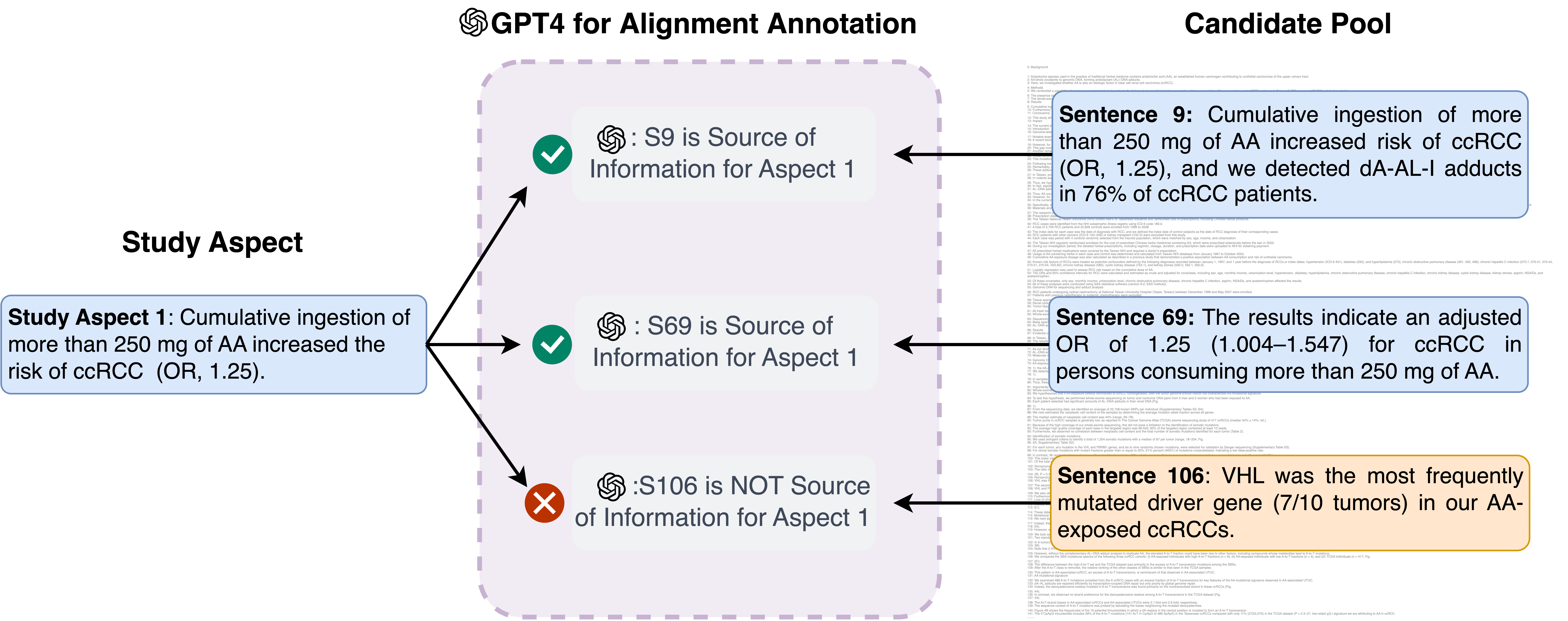}
    \caption{\small{The process for matching sentences with study aspects. In this example, GPT4 sees the study aspect, one candidate sentence from the research paper with the context, and is prompted to determine whether the candidate is a source of information for the study aspect.
    }}
    \label{fig:annotation}
\end{figure}

For the annotation pipeline, GPT-4 is shown a target sentence from the research paper and a study aspect (as well as some additional context: the 10 surrounding sentences from the research paper, and the evidence summary from the review paper). It is then asked to evaluate whether the target sentence implies most of the information contained in the study aspect. See Figure \ref{fig:annotation}.

The optimization and evaluation of the pipeline were performed using a development set/test set split. The prompt and annotation methodology were optimized on a development set of 37 research papers. The prompt and optimization procedure is provided in Appendix \ref{sec:appendix_automatic alignment annotation procedure}.

After the pipeline was finalized, it was evaluated on a test set of 50 randomly sampled research papers. For each research paper, a single study aspect was selected, and every sentence in the paper was annotated for this study aspect. The labeling was performed by four annotators, who are Ph.D. researchers in bioinformatics. The annotators were split into two teams. Each annotator first performed the annotation task independently. The pairs within each team then consulted with each other to reach consensus judgments. Finally, inter-annotator agreement was calculated by comparing the judgments of the two teams. Full annotation guidelines are provided in Appendix \ref{sec:appendix_annotation guideline}.

Each team labeled 8111 (sentence, study aspect) pairs in total. Table \ref{tab:hypothesis_test} shows inter-annotator agreement between the two human teams, and between the human teams and GPT-4. Human and GPT-4 judgments match each other more than 98\% of the time. Because of class imbalance (positive labels are rare, around 150 out of 8111), other measures of agreement such as Cohen's $\kappa$ are in the mid 60's, indicating substantial agreement between the human teams and between the human teams and GPT-4. Bootstrapped hypothesis tests find no significant difference between the human/human agreement rate and the human/GPT-4 agreement rate. To account for potential correlations among annotations from the same research paper, we additionally computed confidence intervals using hierarchical bootstrap sampling, and found similar results, shown in Appendix \ref{sec:appendix_hypothesis testing results}. 

For the larger EvidenceBench-100k, over 150 million sentence-aspect pair judgements would need to be made. We switch the annotator to GPT-4o-mini-0718. We validated the quality of its annotation by re-running the hypothesis test on the same 50 randomly sampled papers, and found very similar Human \& GPT agreement. See Appendix \ref{sec:appendix_hypothesis testing results} for 4o-mini's results.

\begin{table}[htbp]
    \caption{\small{Hypothesis Testing Results for Automatic Alignment Annotation}}

    \label{tab:hypothesis_test}

    \centering
    \setlength{\tabcolsep}{4pt}
    \begin{tabular}{lcccccc}

        \toprule

        \textbf{Metrics} & \textbf{Human \& Human Average} & \textbf{Human \& GPT Average} & \textbf{p-value} \\

        \midrule

        Exact Accuracy & 98.8 $\pm$ 0.3 & 98.7 $\pm$ 0.2 & 0.21 \greencheck \\

        F1 Binary & 66.0 $\pm$ 6.5  & 64.6 $\pm$ 5.6 & 0.64 \greencheck \\

        Cohen's $\kappa$ & 65.4 $\pm$ 6.6 & 63.9 $\pm$ 5.6  & 0.63 \greencheck\\

        Spearman's $\rho$ & 65.4 $\pm$ 6.6 & 64.0 $\pm$ 5.6 & 0.65 \greencheck\\

        \bottomrule

    \end{tabular}

\end{table}

\section{Experiment Setup}
\label{sec:experiment}

\textbf{EvidenceBench Tasks:} we consider four evidence retrieval tasks with different settings.

\textbf{Evidence Retrieval @Optimal:} denoted as ER@optimal. The smallest number of sentences to cover all aspects is denoted as Optimal. The average optimal number is only 4.6. The task is for a model to retrieve no more than the Optimal number of sentences to form the best evidence set relevant to the hypothesis. \textbf{Evidence Retrieval @10} is denoted as ER@10. See Appendix \ref{ssec:appendix_default_prompt_eroptimal_10} for the prompt.

\textbf{Result Evidence Retrieval @Optimal:} denoted as Result-ER@Optimal. The task restricts the model to retrieve no more than the optimal number of sentences, which is calculated by the minimum number of sentences required to cover all study aspects labeled as "Results". This labeling is done by GPT4-0125. Empirically, half of the aspects are labeled as "Results', see Table \ref{tab:testset statistics}. \textbf{Result Evidence Retrieval @5} is denoted as Result-ER@5. See Appendix \ref{ssec:appendix_prompt_results_eroptimal_5} for the prompt.

\textbf{Experiment Models} We test the following models Claude3-Opus \citep{claude3}, Gemini 1.5 \citep{gemini_15}, GPT-4o \citep{gpt_4o}, Llama3-70B, Llama3-8B \citep{llama3}, E5-v2 \citep{e5}, OpenAI Embedding v3 \citep{openai_text_embedding_3}, VoyageAI v2 \citep{voyage}, GritLM-7B \citep{gritLM}, E5-Mistral-7B \citep{e5_mistral}, NV-Embed-v2\citep{nv-embed} which is the leader of the Massive Text Embedding Benchmark MTEB \citep{mteb}.

\subsection{Evaluation strategies}
\label{ssec:evaluation strategies}
We implement three standard practices for evaluating LLMs on long-context benchmarks \citep{longbench, infty_bench}. \textbf{Chain-of-Thought (CoT):} our default evaluation strategy, optimized from our train set. \textbf{In-Context-Learning (ICL):} From the train set, we randomly sample 8 example hypotheses and their corresponding ground-truth set of retrieved sentences, as per standard ICL practices \citep{icl1, icl2}, in addition to our default CoT prompt. \textbf{Section-by-Section (Sec-by-Sec):} We divide a research paper by its natural sections. In the first stage, LLM only retrieves from each section one at a time. In the second stage, all retrieved sentences are presented to the LLM and final selections are made. 

Our evaluation metric is Aspect Recall, defined in Section \ref{ssec:evaluation metics}. In Appendix \ref{sec:appendix_experiment details}, we presented all prompts used in our experimental evaluation for LLMs, instructions for embedding models as well as standard errors for model evaluations.

\begin{table}[htbp]
    \caption{\small{Four tasks are reported on EvidenceBench test set. For each model, the highest number is reported if multiple strategies are used.}}
    \label{tab:comparison}
    \centering
    \small
    \setlength{\tabcolsep}{2pt}
    \begin{tabular}{l|cccc|ccccc}
        \toprule
         & GPT-4o & Claude3 & Gemini & LLama3-70B & OpenAI & Voyage & GritLM & E5-Mistral&NV-Embed \\
        \midrule
        ER@Optimal  & \textbf{51.4} & 47.6 & 48.3 & 46.7 & 25.1 & 22.7 & 27.0 & 22.7 & 25.2\\
        ER@10 & \textbf{71.6} & 66.4 & 65.4 & 65.4 & 42.2 & 42.0 & 46.4 & 41.9 & 44.7 \\
        \midrule
        Result-ER@Opt & \textbf{52.6} & 51.7 & 46.7 & 46.2 & 19.1 & 18.3 & 18.9 & 19.3 & 20.1\\
        Result-ER@5  & \textbf{70.8} & 68.7 & 65.4 & 63.7 & 33.1 & 31.9 & 39.1 & 33.6 & 35.6\\
        \bottomrule
    \end{tabular}

\end{table}

\section{Results}
\label{sec:results}

From Table \ref{tab:comparison}, we list the best performance for each model (LLM or embedding model) on the four constrained Evidence Retrieval tasks in the original EvidenceBench. GPT-4o consistently outperforms others across all tasks, while Gemini, Claude3-Opus, and Llama3-70B closely trail behind. Llama3-70B can only be evaluated using the Sec-by-Sec strategy due to its limited context window, but it shows robust performance across all tasks. There is a qualitative difference between LLMs and embedding models, partially because embedding models are not context-aware when calculating sentence embeddings. This invites future work on general-purposed context-aware embedding models.

\textbf{In-context learning:} From Table \ref{tab:gen_models}, we see an 8-shot ICL does not significantly alter performances for LLMs. In particular, GPT-4o and Gemini-1.5 slightly improve, while Claude3-Opus slightly degrades. This indicates the primary difficulty is context-length and not a failure to understand task requirement, suggesting that ICL is less effective on long-context benchmarks.

\textbf{Section-by-Section Processing:} On the other hand, from Table \ref{tab:gen_models}, Sec-by-Sec considerably improves Gemini and Claude's performances, suggesting that the default longer-context version of the task hinders their retrieval abilities. Section-by-section is by far the most robust strategy observed here. We performed further analyses in Appendix \ref{sec:analysis}.

\textbf{Reasoning-based LLMs:} We tested two reasoning-based LLMs, Gemini 2.5 Pro \citep{gemini_2_5_pro} and Claude 4 Sonnet \citep{claude4}, using the baseline prompt for ER@Optimal. As we see in Table \ref{tab:reasoning_based models using baseline prompt}, they scored 52.5 and 49.1 on ER@Optimal, respectively, numerically higher than GPT-4o's performance. We leave a more extensive evaluation of reasoning models on EvidenceBench for future work.
\begin{table}[ht]
    \centering
        \centering
        \caption{\small{Comparison of different strategies for LLMs on the EvidenceBench test set. 
        }
        }
        \label{tab:gen_models}
         \setlength{\tabcolsep}{2pt}
\begin{tabular}{l>{\centering\arraybackslash}m{1.5cm}>{\centering\arraybackslash}m{1.5cm}|>{\centering\arraybackslash}m{1.5cm}>{\centering\arraybackslash}m{1.5cm}|>{\centering\arraybackslash}m{1.5cm}>{\centering\arraybackslash}m{1.5cm}}
            \toprule
             & \multicolumn{2}{c}{\textbf{Baseline}} & \multicolumn{2}{c}{\textbf{ICL}} & \multicolumn{2}{c}{\textbf{Sec-by-Sec}} \\
            \cmidrule(lr){2-3} \cmidrule(lr){4-5}  \cmidrule(lr){6-7}
            \textbf{Model} & ER 
            @
            
            Optimal & ER@
            
            10 & ER 
            @
            
           Optimal & ER@
            
            10 & ER 
            @
            
            Optimal & ER@
            
            10 \\
            \midrule
            GPT-4o & \textbf{48.1} & \textbf{69.6} & \textbf{51.4} & \textbf{68.7} & \textbf{50.9} & \textbf{71.6} \\
            Claude3 & 41.1 & 53.6 & 38.3 & 55.4 & 47.6 & 66.4 \\
            Gemini & 42.7 & 63.0 & 43.2 & 62.4 & 48.3 & 65.4 \\
            Llama3-70B & - & - & - & - & 46.7 & 65.4 \\
            \bottomrule
        \end{tabular}

\end{table}

\begin{table}[htbp]
    \caption{\small{Evaluation for reasoning-based LLMs using baseline prompt}}
    \label{tab:reasoning_based models using baseline prompt}
    \centering
    \small
    \setlength{\tabcolsep}{4pt}
    \begin{tabular}{l|ccccc}
        \toprule
        Model & Gemini 2.5 Pro & Claude 4 Sonnet & GPT-4o & Claude 3 Opus & Gemini 1.5 \\
        \midrule
        ER@Optimal & \textbf{52.5} & 49.1 & 48.1 & 41.1 & 42.7 \\
        \bottomrule
    \end{tabular}
\end{table}

\subsection{Fine-tuning and Evaluation on EvidenceBench-100k}

EvidenceBench-100k is split into a 80k train set and a 20k test set. For cost reasons, we randomly sample 3000 datapoints from the 20k test set to evaluate. Table \ref{table:new_test_set} shows EvidenceBench-100k test set can be used to evaluate and clearly differentiate various models' performance.

Furthermore, we fine-tune two models: E5-v2 335M and Llama3-8B (sec-by-sec strategy) using the 80k training datapoints. We test them on the original EvidenceBench test set for the task of Result-ER@Optimal. We notice both fine-tuned models show significant improvements over their baselines as shown in Table \ref{table:fine_tuned}. This shows the EvidenceBench-100k train set can be used for model developments. See full details of fine-tuning at Appendix \ref{appendix:finetuning}.

\begin{table}[h!]
\caption{\small{Comparison of Model Performances on EvidenceBench Datasets}}
\centering
\begin{subtable}[t]{0.48\textwidth}
\caption{EvidenceBench-100k test set}
\centering
\begin{tabular}{l|c}
\hline
\rule{0pt}{2.5ex}\textbf{Model} & \textbf{Result-ER@Optimal} \\
\hline
\rule{0pt}{2.5ex}GPT-4o & \textbf{42.84}\% \\
\rule{0pt}{2.5ex}Claude3 & 35.12\% \\
\rule{0pt}{2.5ex}GritLM & 14.59\% \\
\rule{0pt}{2.5ex}OpenAI & 10.95\% \\
\hline
\end{tabular}

\label{table:new_test_set}
\end{subtable}%
\hfill
\begin{subtable}[t]{0.5\textwidth}
\caption{EvidenceBench test set.}
\centering
\begin{tabular}{l|c}
\hline
\rule{0pt}{2.5ex}\textbf{Model} & \textbf{Result-ER@Optimal} \\
\hline
\rule{0pt}{2.5ex}Pretrained Llama3-8B & 35.8\% \\
\rule{0pt}{2.5ex}Finetuned Llama3-8B & \textbf{41.0}\% \\
\rule{0pt}{2.5ex}Pretrained E5-v2 & 15.2\% \\
\rule{0pt}{2.5ex}Finetuned E5-v2 & \textbf{32.9}\% \\
\hline
\end{tabular}

\label{table:fine_tuned}
\end{subtable}

\label{table:side_by_side}
\end{table}

\subsection{Effect of Hypothesis Negation}
\label{ssec:negation}

In EvidenceBench, most of the sentences which are annotated as relevant to a hypothesis provide evidence in support of the hypothesis, rather than against it. To evaluate whether models are able to identify information which provides evidence against a hypothesis, we created a negated version of EvidenceBench by negating each hypothesis. See Appendix \ref{ssec:appendix_negation prompt} for the prompt and some examples of negated hypotheses. By negating each hypothesis, we have transformed the support relationships into refutation relationships.

Using the baseline prompt, we evaluated GPT-4o and Claude3-Opus on this negated version of the original EvidenceBench test set. In Table \ref{tab:gen_models}, the ER@Optimal for GPT-4o is 48.1 on the original set, the new experiment shows the ER@Optimal drops to 46.7 on the negated set. The ER@Optimal for Claude3 is 41.1 on the original set and the ER@Optimal drops to 36.7 on the negated set. The results demonstrate that both models show decreased performance when evaluating negated hypotheses, indicating that the task becomes more challenging when models must identify evidence that refutes rather than supports a claim. However, the decrease in performance is small for GPT-4o.

\section{Related Work}
\label{sec:related work}

\textbf{Hypothesis Generation}
Recent works explore using LLMs to generate scientific hypotheses \citep{latent_space_bioelectricity, psych_hypothesis_ai, can_chatgpt, researchagent,lab_validation_breast_cancer}. \citet{hypothesis_proposers} fine-tune LLMs on biomedical literature that pairs background knowledge with corresponding hypotheses, and then use the LLMs to generate hypotheses when prompted with background knowledge. 

\textbf{Evidence Retrieval}
Claim-based retrieval \citep{complex_claim} retrieve evidence by breaking down a complex claim into specific aspects and retrieving each aspect. On the other hand, our pipeline uses a novel approach, by decomposing summarized evidence from review papers into study aspects (instead of claims), which serves as ground-truth human domain experts knowledge that would guide our LLM-annotator to match sentences from research papers to these study aspects.

\textbf{LLM in Biomedicine}
Researchers have shown strong performance of LLMs in BioNLP tasks, including relation extraction, question answering, document classification, name entity recognition, and summarization \citep{BioGPT, llm_in_bio, llm_bio_NER, bio_eval, eval_llm_bio, bioNER}. LLMs are also being used to extract specific information from report, e.g. Interventions, Outcomes, and Findings by \cite{extractICO}.

\section{Conclusion}
We introduced EvidenceBench and EvidenceBench-100k, a benchmark for retrieving evidence for scientific hypotheses from biomedical literature. EvidenceBench was constructed using an automated, scalable pipeline that transforms expert-written summaries into fine-grained annotations linked to specific sentences in research papers.

\newpage

\clearpage
\bibliography{colm2025_conference}
\bibliographystyle{colm2025_conference}

\clearpage
\appendix

\textbf{Appendix: Table of Contents}

\begin{enumerate}[label=\Alph*.]
    \item Dataset\dotfill \pageref{sec:appendix_dataset}
    \begin{enumerate}[label=\arabic*.]
        \item Dataset License and Code License\dotfill \pageref{ssec:appendix_dataset license}
        \item Dataset Hosting, Accessibility and Maintenance\dotfill \pageref{ssec:appendix_dataset hosting, accessibility and maintenance}
        \item Dataset Coverage Analysis\dotfill\pageref{ssec:appendix_dataset coverage analysis}
        \item A Motivating Example\dotfill \pageref{ssec:appendix_a motivating example}
        \item Dataset Collection and Processing\dotfill \pageref{ssec:appendix_data_collection_processing}
        \item EvidenceBench Dataset Structure\dotfill \pageref{ssec:appendix_dataset_structure}
    \end{enumerate}
    \item Annotation Guidelines\dotfill \pageref{sec:appendix_annotation guideline}
    \begin{enumerate}[label=\arabic*.]
    \item Guidelines for Hypothesis Validation\dotfill \pageref{ssec:appendix_guideline_expert_verification_hypotheses}
    \item Annotation Guidelines for Alignment of Study Aspects and Sentences\dotfill \pageref{ssec: appendix_guideline_Alignment of Aspects and Sentences}
    \begin{enumerate}[label=\arabic*.]
    \item First Annotation Stage\dotfill \pageref{sssec: appendix_First Stage, Independent Stage}
    \item Second Annotation Stage\dotfill \pageref{sssec:appendix_Second Stage, Collaborative Stage}
    \end{enumerate}
    \end{enumerate}
    \item Automated Alignment Procedure\dotfill \pageref{sec:appendix_automatic alignment annotation procedure}
    \begin{enumerate}[label=\arabic*.]
    \item Study Aspect Decomposition\dotfill \pageref{ssec:appendix_Study Aspect Decomposition from IARC Summary}
    \end{enumerate}
    \item Hypothesis Testing Results\dotfill \pageref{sec:appendix_hypothesis testing results}
    \item Experiment Details\dotfill \pageref{sec:appendix_experiment details}
    \begin{enumerate}[label= \arabic*.]
        \item Default Prompt Template for Evidence Retrieval\dotfill \pageref{ssec:appendix_default_prompt_eroptimal_10}
        \item Prompt Template for Results Evidence Retrieval @Optimal and @5\dotfill \pageref{ssec:appendix_prompt_results_eroptimal_5}
        \item ICL Prompt for Evidence Retrieval @Optimal and @10\dotfill \pageref{ssec:appendix_icl_prompt_eroptimal_10}
        \item Section-by-Section Prompt for Evidence Retrieval @Optimal and @10\dotfill \pageref{ssec:appendix_secbysec_eroptimal_10}
        \item Regeneration Prompt\dotfill \pageref{ssec:appendix_regeneration prompt}
        \item Instructions for Embedding Model\dotfill \pageref{ssec:appendix_instruction_embedding}
        \item Standard Errors for Model Evaluations\dotfill \pageref{ssec:appendix_standard_error in model evaluation}
        \item Negation Prompt\dotfill \pageref{ssec:appendix_negation prompt}
    \end{enumerate}
    \item Model Sensitivity to Paraphrased Hypothesis\dotfill
    \pageref{sec:appendix_model sensitivity to hypothesis}
    \item Hypothesis Collection\dotfill \pageref{sec:appendix_hypothesis generation}
    \item Data Preprocessing\dotfill \pageref{sec:appendix data preprocessing}
    \begin{enumerate}[label= \arabic*.]
        \item Harvesting Evidence Summary\dotfill \pageref{ssec:appendix harvesting evidence summary}
        \item Further Processing Evidence Summary\dotfill \pageref{ssec:appendix_extracting evidence summary}
        \item Identifying Suitable Evidence Summaries\dotfill \pageref{ssec:appendix_identifying suitable evidence summary}
        \item Human Verification\dotfill \pageref{ssec:human verification of harvest}
       \end{enumerate}
    \item Fine-tuning and Evaluation on EvidenceBench-100k\dotfill \pageref{appendix:finetuning}
    \item Qualitative Analysis for GPT-4o on the Original EvidenceBench\dotfill \pageref{appendix:qualitative analysis for gpt-4o}
    \item Further Analyses\dotfill\pageref{sec:analysis}
\end{enumerate}

\clearpage

\section{Dataset}
\label{sec:appendix_dataset}

\subsection{Dataset License and Code License}
\label{ssec:appendix_dataset license}
The EvidenceBench dataset uses the following licenses:
\begin{itemize}
    \item Test set: Provided under CC-BY license.
    \item Train set: Provided under CC-BY-NC-SA license.
    \item Dev set: Provided under CC-BY-NC-SA license.

The EvidenceBench-100 dataset uses CC-BY license.

\end{itemize}
A copy of the full license can be found at \href{https://github.com/EvidenceBench/EvidenceBench/blob/main/LICENSE.md}{https://github.com/EvidenceBench/EvidenceBench/blob/main/LICENSE.md}. Note that the test set has the most permissive license.

All code is released under the MIT License. 
The full license can be found at
\href{https://github.com/EvidenceBench/EvidenceBench/blob/main/LICENSE.md}{https://github.com/EvidenceBench/EvidenceBench/blob/main/LICENSE.md}.

\subsection{Dataset Hosting, Accessibility and Maintenance}
\label{ssec:appendix_dataset hosting, accessibility and maintenance}

The EvidenceBench and EvidenceBench-100k datasets can be accessed at (\href{https://github.com/EvidenceBench/EvidenceBench/}{https://github.com/EvidenceBench/EvidenceBench}).

\subsection{Dataset Coverage Analysis}
\label{ssec:appendix_dataset coverage analysis}
In this paper, we introduce EvidenceBench with 426 papers and EvidenceBench-100k with 107,461 papers. Our two datasets are fully open-sourced under the CC-BY license and encompass a comprehensive range of biomedical topics. Specifically, we quantified topical diversity for the 100k papers using 2025 MeSH headings to demonstrate representativeness. The dataset covers all 16 top-level MeSH branches (A–N, V, Z), confirming coverage of every major biomedical domain. With 17,505 distinct MeSH descriptors representing 56.5\% of the entire 2025 vocabulary (30,956 terms), the breadth matches that of a full-year MEDLINE slice. The branch-level distribution shows a Gini-Simpson diversity of 0.81 (where 1 indicates perfect evenness), demonstrating that no single area dominates the corpus. Additionally, MeSH tree numbers reach a median depth of 5 and 90th-percentile of 7, indicating the benchmark encompasses both fine-grained molecular and procedural topics alongside high-level concepts.

\subsection{A Motivating Example}
\label{ssec:appendix_a motivating example}
Aristolochic Acid (AA) is a toxin that is naturally occurring in traditional Chinese herbal medicines and has been known to cause many types of cancer in animals and humans \cite{aristolochic_acid}. However, 20 years ago, the causal relationship between AA and kidney cancer was not yet confirmed. In this section, we present an example data instance related to AA.

\textbf{Hypothesis:}

Aristolochic Acid (AA) induced DNA mutation is causal for renal carcinoma (RCC).

\textbf{Evidence Summary} 

Results from \cite{aristolochic_acid} showed that a cumulative ingestion of more than 250 mg of AA increased the risk of ccRCC with an odds ratio (OR) of 1.25. A distinctive AA mutational signature was evident in 6/10 sequenced ccRCC exomes from AA-exposed patients. Among these tumors, VHL, the most frequently mutated gene, mutated in 7 out of 10 samples.

\textbf{Study Aspect Decomposition:}
\begin{enumerate}
    \item Cumulative ingestion of more than 250 mg of AA increased the risk of ccRCC  (OR, 1.25). [Sentences 9 and 69 are sources of information for Aspect 1].
    \item A distinctive AA mutational signature was evident in six of the 10 sequenced ccRCC exomes from AA-exposed patients. [Sentence 163 is the source of information for Aspect 2]
    \item The most frequently mutated driver gene was VHL. [Sentence 106 is the source of information for Aspect 3].
    \item VHL was mutated in 7 out of 10 tumors. [Sentence 106 is the source of information for Aspect 4]
\end{enumerate}

\textbf{Full Paper:}

All 216 sentences from  \cite{aristolochic_acid} are indexed, starting from 0 to 215. For brevity, we will not reproduce the entire paper here. Figure \ref{fig:evaluation} shows sentences 9, 69, 106 from the full list of sentences are retrieved by a model.

\textbf{Selected Sentences from Full Paper:}
\begin{itemize}
    \item Sentence 9: Cumulative ingestion of more than 250 mg of AA increased risk of ccRCC (OR, 1.25), and we detected dA-AL-I adducts in 76\% of Taiwanese ccRCC patients.
    \item Sentence 69: The results (Table 1) indicate an adjusted OR of 1.25 (1.004–1.547) for ccRCC in persons consuming more than 250 mg of AA during the period of 1997 to 2003.
    \item Sentence 106: VHL was the most frequently mutated driver gene (7/10 tumors) in our AA-exposed ccRCCs (Table 2).
    \item Sentence 163:  Whole-exome sequencing confirmed that the AA mutational signature was present in 6 of 10 ccRCC patients studied.
\end{itemize}

\subsection{Dataset Collection and Processing}
\label{ssec:appendix_data_collection_processing}
We use BioC  API to download biomedical papers which are available in the PMC database. Papers unavailable in PMC are manually downloaded. We use GROBID  to parse papers from PDF format to XML format. We use Stanza  to split paragraphs of text into sentences. We manually copied and pasted all required open-access review paper sections, and do not distribute any contents of these review papers. 

\subsection{EvidenceBench and EvidenceBench-100k Datasets Structure}
\label{ssec:appendix_dataset_structure}
EvidenceBench uses a train, dev, test split. EvidenceBench-100k uses a train and test split. All datasets have the same structure.
Each data instance (stored in a JSON) has the following features:
\begin{enumerate}[label=-]
    \item \texttt{hypothesis}: the biomedical hypothesis in string format. 
    \item \texttt{paper\_as\_candidate\_pool}:  an ordered tuple of strings. Each string is one sentence from the paper. This serves as the candidate pool for all of the evidence retrieval tasks.
    \item \texttt{aspect\_list\_ids}: a list of strings. Each string is an id for a study aspect. 
    \item \texttt{results\_aspect\_list\_ids}: a list of strings. Each string is an id for an aspect related to the study's results.
    \item \texttt{aspect2sentence\_indices}: a mapping (i.e. dictionary) from each aspect to all sentence indices that are sources of information for that aspect. 
    \item \texttt{sentence\_index2aspects}: a mapping (i.e. dictionary) from each sentence index to all aspects that this sentence is a source of information for.
    \item \texttt{evidence\_retrieval\_at\_optimal\_evaluation}: A dictionary that contains information for evaluating a model's performance on the task Evidence Retrieval @Optimal.
    \begin{enumerate}[label={\textbullet}]
        \item \texttt{optimal}: A positive integer, which is the smallest number of sentences needed to cover all study aspects. 
        \item \texttt{one\_selection\_of\_sentences}: a list of sentence indices, containing the smallest number of sentences needed to cover all aspects. Note, there are potentially other lists of sentences of the same size which cover all aspects. 
        \item \texttt{covered\_aspects}: the list of aspects that are covered, which is all aspects in this case. 
    \end{enumerate}
    \item \texttt{evidence\_retrieval\_at\_10\_evaluation}: A dictionary that contains information for evaluating a model's performance on the task Evidence Retrieval @10.
    \begin{enumerate}[label={\textbullet}]
        \item \texttt{one\_selection\_of\_sentences}: a list of 10 sentence indices. This list covers the maximum number of aspects which can be covered by 10 sentences.
        \item \texttt{covered\_aspects}: the list of aspects that are covered, which may be fewer than all aspects.
    \end{enumerate}
    \item \texttt{results\_evidence\_retrieval\_at\_optimal\_evaluation}: A dictionary that contains the information for evaluating a model's performance on the task Results Evidence Retrieval @Optimal. The structure is similar to \texttt{evidence\_retrieval\_at\_optimal\_evaluation}.
    \item \texttt{results\_evidence\_retrieval\_at\_5\_evaluation}: A dictionary that contains the necessary information for evaluating a model's performance on the task Results Evidence Retrieval @5. The structure is similar to \texttt{evidence\_retrieval\_at\_10\_evaluation}.
    \item \texttt{sentence\_types\_in\_candidate\_pool}: a tuple of strings. Each string is a sentence type. There are three possible sentence types: section\_name, abstract, and normal\_paragraph. For example, if the third string is 'abstract', that means the third sentence comes from the abstract.
    \item \texttt{paper\_id}: the id of the paper used as the candidate pool.
\end{enumerate}

\section{Annotation Guidelines}
\label{sec:appendix_annotation guideline}
\subsection{Guidelines for Hypothesis Validation}
\label{ssec:appendix_guideline_expert_verification_hypotheses}
Below, we show the annotation guidelines for evaluating the hypotheses extracted from the review papers. These guidelines were co-designed and approved by a medical doctor who did not see the 50 hypotheses which were evaluated. \\

\texttt{\textbf{Overall}:}\\

\texttt{IARC is a WHO organization that invites field experts to write a review about the potential carcinogenicity of a certain chemical/compound/product/substance, where they survey many relevant papers.}

\texttt{A review is typically organized into the following sections:}

\begin{itemize}
    \item \texttt{Exposure Data (e.g., how humans and animals come into contact with the substance).} 
    \item \texttt{Animal Study.} 
    \item \texttt{Human Study.}
    \item \texttt{Mechanistic Evidence (e.g., the mechanism for carcinogenicity).}
    \item \texttt{Others.}
\end{itemize}

\texttt{For each relevant paper, the field experts will extract certain information from the paper, for a specific purpose, which does not have to align with the original goal of the paper.} \\

\texttt{\textbf{Annotation Task:}}\\

\texttt{Each task is in a docx file. In each docx file, you will see:}
\begin{itemize}
    \item \texttt{A hypothesis.}
    \item \texttt{A paragraph of extracted information from paper.}
    \item \texttt{A reference page (For reference only).}
    \begin{itemize}
        \item \texttt{Potentially more context for the hypothesis (i.e., a potential connection between the hypothesis and the extracted information from paper.}
        \item \texttt{The review that contains the extracted information from the paper.}
        \item \texttt{Link for the paper.}
    \end{itemize}
\end{itemize}
\texttt{You have two tasks.}\\
\begin{itemize}
    \item \texttt{Determine if the hypothesis is a reasonable hypothesis, given your understanding of the hypothesis and your external knowledge and experience.}
    \begin{itemize}
        \item \texttt{The hypothesis might be about the carcinogenicity of a substance (for human or animal), or might be about how humans get exposed to a substance, or might be about experimental procedure, or something else.}
        \item \texttt{Determine if the hypothesis is a valid statement with scientific value, it could be a false statement, but disproving it would have scientific value. In other words, you should not judge the accuracy of the hypothesis. You should only judge if the hypothesis contains scientific value.}
        \item \texttt{Determine if the hypothesis looks like a hypothesis, i.e., has the format of a real hypothesis.}
        \item \texttt{Make your judgment based only on the contents of the hypothesis, which is usually just one sentence. Your decision should not be influenced by the other task or other materials you see, though for better comprehension, you can refer to the links on the reference page.}
        \item \texttt{Record your decision(Yes or No), and leave any optional comment if you want. If you think the answer is not binary, then you do not have to write yes or no, but you have to give an explanation.}
    \end{itemize}
    \item \texttt{Determine if the extracted information from the paper contains evidence that can potentially help support or refute the hypothesis.}
    \begin{itemize}
        \item \texttt{Answer Yes or No, followed by a brief explanation. One or two sentences. If you think the answer is not binary, then you do not have to write yes or no, but you have to give an explanation.} 
    \end{itemize}
\end{itemize}

\texttt{\textbf{Notes:}}\\

\texttt{You have to pledge the following conditions are met during annotation for each task packet.}
\begin{itemize}
    \item \texttt{No consulting with AI and LLM. }
    \item \texttt{For words or concepts that you are not familiar with and believe are important for comprehension, search for them and understand their meaning.}
    \item \texttt{If you do not understand the hypothesis or the extracted information from the paper, you should read the IARC review to better understand the hypothesis and read the full paper to better understand the concepts mentioned in the hypothesis and extracted information from the paper.}
    \item \texttt{You are not required to read the whole paper nor the full IARC review, just to the point when you believe you understand the hypothesis and extracted information from the paper well enough.} 
\end{itemize}

\subsection{Annotation Guidelines for Alignment of Study Aspects and Sentences}
\label{ssec: appendix_guideline_Alignment of Aspects and Sentences}
\subsubsection{First Annotation Stage}
\label{sssec: appendix_First Stage, Independent Stage}
This section shows the annotation guidelines for the first stage of aspect-sentence alignment. In this stage, each annotator had to independently annotate 50 papers.

\texttt{You have a total of 50 annotation task packets. Each task packet is a docx. file that contains the following information.}

\begin{itemize}
    \item \texttt{An aspect (one piece of important information/detail).}
    \item \texttt{The context for the Aspect (summary or a collection of extracted details from a paper).}
    \item \texttt{The URL for the paper (pmc or pubmed link).}
    \item \texttt{The list of indexed text elements of the paper (a text element could be a sentence or a section title).}
\end{itemize}

\texttt{You have to pledge the following conditions are met during annotation for each task packet.} 
\begin{itemize}
    \item \texttt{No consulting with AI or LLM.}
    \item \texttt{For words you are not familiar with and believe are important for comprehension, conduct a search and understand its meaning.}
    \item \texttt{Click on the paper URL and find full contents either in HTML, XML, or PDF format, and read through it from start to finish, at least once.}
    \item \texttt{For every text element in the list, you must look at it and read it at least once. }
    \item \texttt{You cannot talk to other annotators about anything related to your task, including progress and insights.}
    \item \texttt{You have to take a mandatory 5-minute break after every 1 hour of performing annotation. }
    \item \texttt{You cannot exceed 8 hours of annotation per day. }
\end{itemize}

\texttt{Below is the recommended procedure for annotating each packet.}

\begin{itemize}
    \item \texttt{Read and understand the aspect, and the context of the aspect.}
    \item \texttt{Decide what details in the aspect count as information, and what details count as context. At least one detail needs to be identified as information, but an aspect can do without context if it is self-contained and very clear. }
    \begin{itemize}
        \item \texttt{You can decide what information is, but geographical, temporal, and numerical data are all information.}
        \item \texttt{Typically, context is recurring and ubiquitous information throughout the paper.}
    \end{itemize}
    \item \texttt{For each text element, you decide if a very significant amount of information identified in the aspect is also explicitly present in the text element. Note, information has to be explicitly present, it cannot be from inference or allusion. Acronyms, abbreviations, and different presentation formats of the same information (e.g., rounding of numbers) are acceptable, as long as it is clear to you. }
    \begin{itemize}
        \item \texttt{Even if you find significant information overlap, you have to make sure the sentence is in the same context as the aspect.}
        \begin{itemize}
            \item \texttt{Same context typically refers to the same study or experiment.}
            \item \texttt{Check if the sentence refers to the same experiment as the aspect, since different experiments could be in one paper.}
        \end{itemize}
    \end{itemize}
    \item \texttt{Do not do complicated mental inference. Once you have a clear understanding of the aspect and sentence, do not try to invent a spurious connection between sentence and aspect. }
    \begin{itemize}
        \item \texttt{Specifically, do not do complex computations of numbers.}
    \end{itemize}
    
\end{itemize}

\subsubsection{Second Annotation Stage}
\label{sssec:appendix_Second Stage, Collaborative Stage}
Below we show the annotation guidelines for the second stage of the aspect-sentence alignment. In this stage, two annotators from the same team come to a consensus on any disagreements from the first stage. 

\begin{itemize}
    \item \texttt{Go through task 0-49.}
    \item \texttt{Resolve your difference, check if you made a mistake, or if you missed something. If you made a conceptual error (e.g. you failed to understand some terminology), you may have to go through the paper again quickly. }
    \begin{itemize}
        \item \texttt{For sentences that you cannot resolve your difference after discussion, i.e., one person says yes, and the other person says no, you should include them as well.}
    \end{itemize}
\end{itemize}

\section{Automated Alignment Procedure}
\label{sec:appendix_automatic alignment annotation procedure}
This section describes prompt optimization for the LLM alignment of study aspects and sentences. 

Prompt optimization was performed with GPT4-0125 on an independent development set of 37 (aspect, paper) pairs, i.e., 37 tasks. There was no overlap with the papers labeled by the two teams of human annotators. 

\begin{figure}[htbp]
    \centering
    \includegraphics[width=\linewidth]{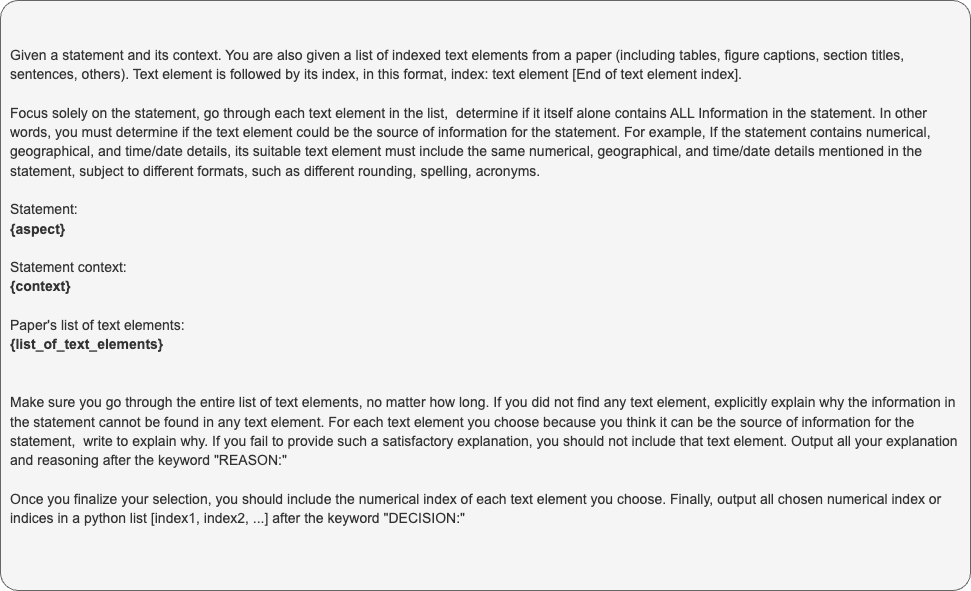}
    \caption{Prompt for aligning a sliding window of consecutive text elements with a study aspect.}
    \label{fig:aspect_sentence_tagging}
\end{figure}
 
In order to reduce the frequency of GPT-4 forgetting information from the papers, we use a sliding window (window-length = 10 sentences) with an overlap of 5 text elements across windows. GPT-4 sees a sliding window of sentences and annotates each sentence according to whether it is a source of information for the aspect. 

Since the sliding windows are overlapping, each text element (except for the first 5) is considered twice by GPT-4. A sentence is labeled as positive if it is selected in either sliding window. 

Figure \ref{fig:aspect_sentence_tagging} shows the final prompt template used for aligning text elements with aspects. Each template uses one aspect, 10 text elements in a sliding window, and the context around the one aspect (i.e., the evidence summary of the paper).  

\newpage
\subsection{Study Aspect Decomposition}
\label{ssec:appendix_Study Aspect Decomposition from IARC Summary}
There are two steps in aspect decomposition. \\
\textbf{The first step} is decomposing a evidence summary into a list of study aspects where each aspect represents a single piece of information. The granularity of the decomposition is determined by the following rule:\\
Each decomposed study aspect must be able to align with at least one sentence from the paper. If a study aspect contains so much information that no one sentence can cover a significant portion of these details, then this study aspect is considered too coarse-grained and must be further decomposed. 

\begin{figure}[htbp]
    \centering
    \includegraphics[width=\linewidth]{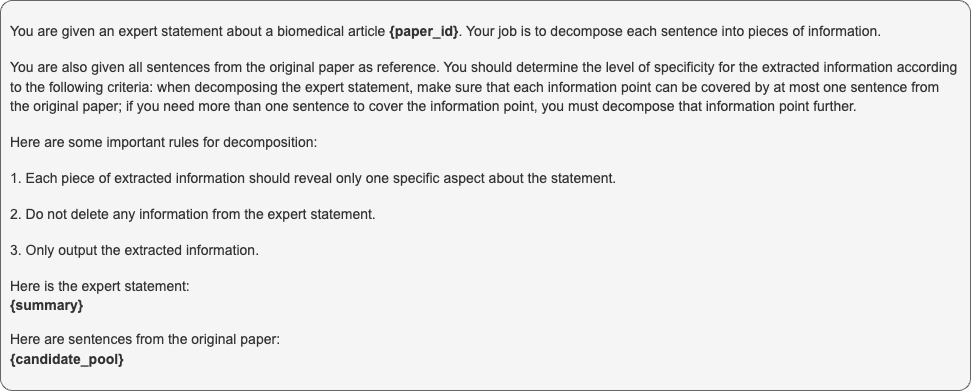}
    \caption{Step 1 of decomposing a evidence summary into aspects, using the granularity condition.}
    \label{fig:aspect_decomposition_stage1}
\end{figure}

See Figure \ref{fig:aspect_decomposition_stage1} for the prompt template that achieves the first step of aspect decomposition.

\begin{figure}[htbp]
    \centering
    \includegraphics[width=\linewidth]{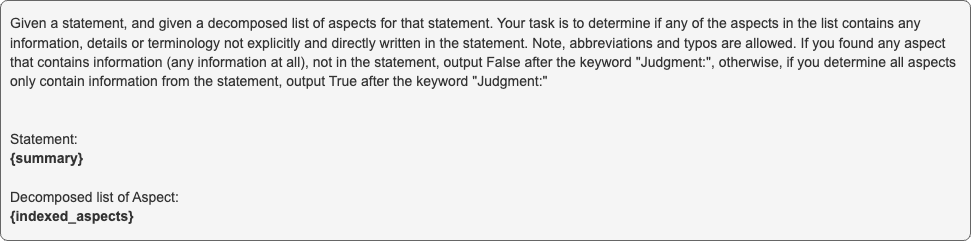}
    \caption{Step 2 of decomposing a evidence summary into aspects. This step confirms that aspects only contain information from the evidence summary.}
    \label{fig:aspect_decomposition_stage2}
\end{figure}

\textbf{The second step} is checking if the decomposed list of study aspects only contains information from the evidence summary, and if no other paper-specific information leaked into the decomposed list. See Figure \ref{fig:aspect_decomposition_stage2} for the prompt template. Any datapoint whose decomposed list of aspects did not pass the second step verification is filtered out. Fewer than 10\% of datapoints are filtered at this step. Empirically we noticed those filtered datapoints have evidence summaries that are not self-contained. Therefore, we did not attempt to recover these datapoints.   

\newpage
\section{Hypothesis Testing Results}
\label{sec:appendix_hypothesis testing results}

During hypothesis testing for automatic alignment annotation, although only 50 papers were annotated, each paper contains roughly 160 (sentence, aspect) pairs, totaling 8,111 pairs. In the paper, correlation and agreement metrics are calculated based on these 8,111 pairs and not on the 50 papers. Because annotations for sentences in the same paper are partially correlated, the effective sample size therefore lies between 50 and 8,111. In order to account for this correlation structure, we can use hierarchical bootstrap sampling (first sampling papers, then sampling sentences within papers) to compute confidence intervals and p-values. We provide results for GPT-4o and GPT-4o-mini on standard bootstrap sampling and hierarchical bootstrap sampling in Tables \ref{tab:hypothesis_test_4o_standard}, \ref{tab:hypothesis_test_4o_hierarchical}, \ref{tab:hypothesis_test_4omini_standard}. \ref{tab:hypothesis_test_4omini_hierarchical}.

\begin{table}[htbp]
    \centering
    \setlength{\tabcolsep}{4pt}
    \begin{tabular}{lcccccc}
        \toprule
        \textbf{Metrics} & \textbf{Human \& Human Average} & \textbf{Human \& GPT Average} & \textbf{p-value} \\
        \midrule
        Exact Accuracy & 98.8 $\pm$ 0.3 & 98.7 $\pm$ 0.2 & 0.21 \greencheck \\
        F1 Binary & 66.0 $\pm$ 6.5  & 64.6 $\pm$ 5.6 & 0.64 \greencheck \\
        Cohen's $\kappa$ & 65.4 $\pm$ 6.6 & 63.9 $\pm$ 5.6  & 0.63 \greencheck\\
        Spearman's $\rho$ & 65.4 $\pm$ 6.6 & 64.0 $\pm$ 5.6 & 0.65 \greencheck\\
        \bottomrule
    \end{tabular}
    \caption{\small{hypothesis testing result for GPT-4o on 8,111 pairs using standard bootstrap sampling}}
    \label{tab:hypothesis_test_4o_standard}
\end{table}

\begin{table}[htbp]
    \centering
    \setlength{\tabcolsep}{4pt}
    \begin{tabular}{lccc}
        \toprule
        \textbf{Metrics} & \textbf{Human \& Human Average} & \textbf{Human \& GPT Average} & \textbf{p-value} \\
        \midrule
        Exact Accuracy & 98.7 $\pm$ 0.5 & 98.6 $\pm$ 0.5 & 0.46 \greencheck\\
        F1 Binary & 66.5 $\pm$ 9.9 & 65.8 $\pm$ 8.8 & 0.91 \greencheck \\
        Cohen's $\kappa$ & 65.8 $\pm$ 10.1 & 65.1 $\pm$ 9.0 & 0.90 \greencheck \\
        Spearman's $\rho$ & 65.9 $\pm$ 10.0 & 65.3 $\pm$ 8.9 & 0.91 \greencheck \\
        \bottomrule
    \end{tabular}
    \caption{\small{Hypothesis testing result for GPT-4o using hierarchical bootstrap sampling.}}
    \label{tab:hypothesis_test_4o_hierarchical}
\end{table}

\begin{table}[htbp]
    \centering
    \setlength{\tabcolsep}{4pt}
    \begin{tabular}{lcccc}
        \toprule
        \textbf{Metrics} & \textbf{Human \& Human Average} & \textbf{Human \& GPT Average} & \textbf{p-value} \\
        \midrule
        Exact Accuracy & 98.7 $\pm$ 0.5 & 98.6 $\pm$ 0.5 & 0.134 \greencheck \\
        F1 Binary & 66.5 $\pm$ 12.3 & 64.0 $\pm$ 9.9 & 0.362 \greencheck \\
        Cohen's $\kappa$ & 65.8 $\pm$ 12.6 & 63.3 $\pm$ 9.9 & 0.356 \greencheck \\
        Spearman's $\rho$ & 65.9 $\pm$ 12.6 & 63.4 $\pm$ 9.9 & 0.360 \greencheck \\
        \bottomrule
    \end{tabular}
    \caption{\small{Hypothesis testing result for GPT-4o-mini on 8,111 pairs using standard bootstrap sampling.}}
    \label{tab:hypothesis_test_4omini_standard}
\end{table}

\begin{table}[h!]
    \centering
    \setlength{\tabcolsep}{4pt}
    \begin{tabular}{lcccc}
        \toprule
        \textbf{Metrics} & \textbf{Human \& Human Average} & \textbf{Human \& GPT Average} & \textbf{p-value} \\
        \midrule
        Exact Accuracy & 98.7 $\pm$ 0.6 & 98.6 $\pm$ 0.6 & 0.339 \greencheck \\
        F1 Binary & 66.5 $\pm$ 9.9 & 64.0 $\pm$ 7.4 & 0.541 \greencheck \\
        Cohen's $\kappa$ & 65.8 $\pm$ 10.1 & 63.3 $\pm$ 7.6 & 0.536 \greencheck \\
        Spearman's $\rho$ & 65.9 $\pm$ 10.0 & 63.4 $\pm$ 7.6 & 0.533 \greencheck \\
        \bottomrule
    \end{tabular}
    \caption{\small{Hypothesis testing result for GPT-4o-mini using hierarchical bootstrap sampling.}}
    \label{tab:hypothesis_test_4omini_hierarchical}
\end{table}

\newpage
\section{Experiment Details}
\label{sec:appendix_experiment details}
There are several prompt templates used for experimental evaluation, which are variations on a default template. \\

\subsection{Default Prompt Template for Evidence Retrieval}
\label{ssec:appendix_default_prompt_eroptimal_10}

\begin{figure}[htbp]
    {\setlength{\abovecaptionskip}{0pt} \setlength{\belowcaptionskip}{0pt}
    \centering
    \includegraphics[width=\linewidth]{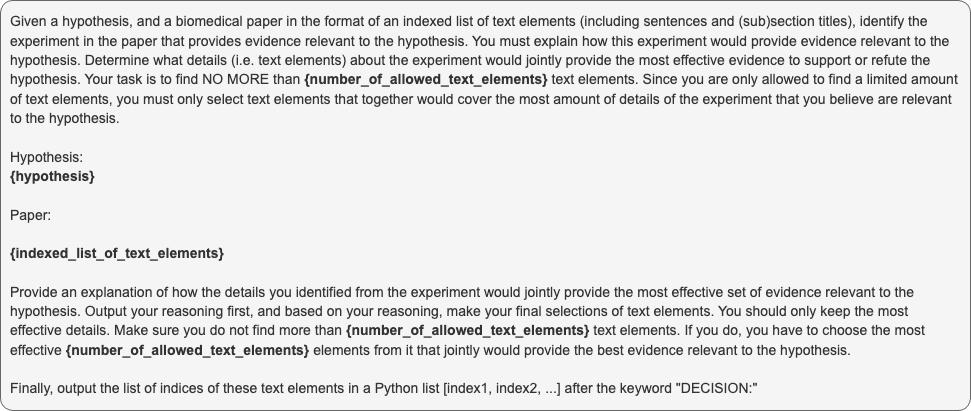}
    \caption{Default prompt template for evaluating LLMs on tasks Evidence Retrieval @Optimal and Evidence Retrieval @10.}
    \label{fig:experiment_baseline}}
\end{figure}

The default prompt template asks an LLM to retrieve no more than K sentences for the Evidence Retrieval tasks.
Figure \ref{fig:experiment_baseline} shows the default prompt template for ER @Optimal or ER @K.

\vspace{-1em}
\subsection{Prompt Template for Results Evidence Retrieval @Optimal or @5}
\label{ssec:appendix_prompt_results_eroptimal_5}
\vspace{-5pt}
\begin{figure}[htbp]
    {\setlength{\abovecaptionskip}{0pt} \setlength{\belowcaptionskip}{0pt}
    \centering
    \includegraphics[width=\linewidth]{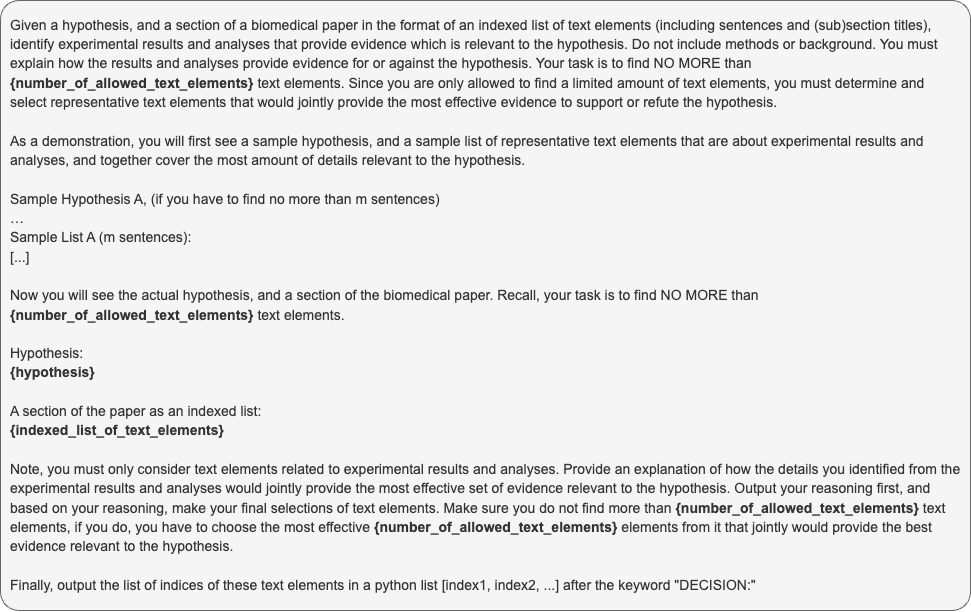}
    \caption{Step 1: Prompt template for evaluating LLMs on tasks \textbf{Results} Evidence Retrieval @Optimal and Evidence Retrieval @10. Here, the number of allowed text elements denotes Optimal or 10. This prompt uses a mixture of one-shot ICL and Section-by-Section.}
    \label{fig:experiment_baseline_result}}
\end{figure}

\newpage
\textbf{Step 1 of Prompt template with one-shot ICL and Section-by-Section}

Figure \ref{fig:experiment_baseline_result} shows the prompt template for any tasks that only focus on retrieving sentences related to experiment results or analyses based on experiment outcomes. Note, this prompt uses a mixture of two strategies: one-shot ICL (in-context-learning) and section-by-section processing. These two strategies are proven effective in the other two tasks, Evidence Retrieval @Optimal and @K. Due to budget limitations, we can only provide this mixture strategy, which proves to be the best strategy on the training set. Note, for fairness, for each section, we can only instruct the LLM to retrieve no more than K sentences, even though a paper could have 10 sections. Consequently, the total number of retrieved sentences for all sections combined sometimes exceed to maximally allowed number of sentences K. Therefore, we have the second step of processing.

\textbf{Step 2 of Prompt template with one-shot ICL and Section-by-Section:}\\

\begin{figure}[htbp]
    \centering
    \includegraphics[width=\linewidth]{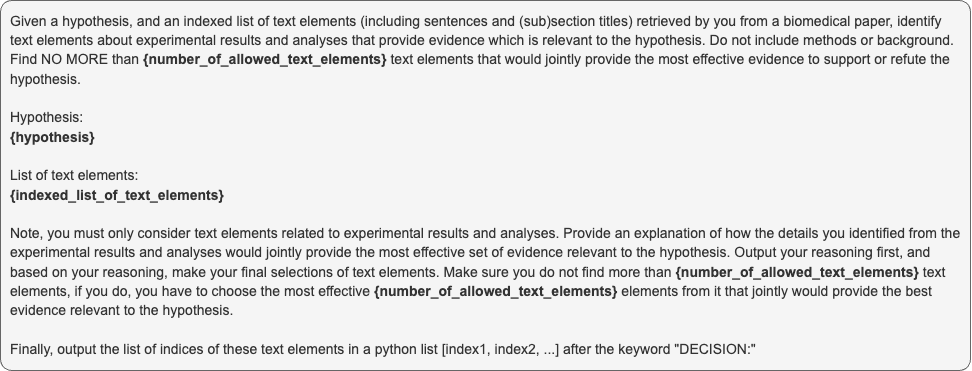}
    \caption{Step 2: Prompt template for evaluating LLMs on tasks \textbf{Results} Evidence Retrieval @Optimal and Evidence Retrieval @10.}
    \label{fig:experiment_baseline_result_step2}
\end{figure}

In step 2, we show the LLM all its retrieved sentences and ask it to select the top K sentences. See Figure \ref{fig:experiment_baseline_result_step2} for its prompt template. 

\newpage
\subsection{ICL Prompt for Evidence Retrieval @Optimal and @10}
\label{ssec:appendix_icl_prompt_eroptimal_10}
\begin{figure}[htbp]
    \centering
    \includegraphics[width=\linewidth]{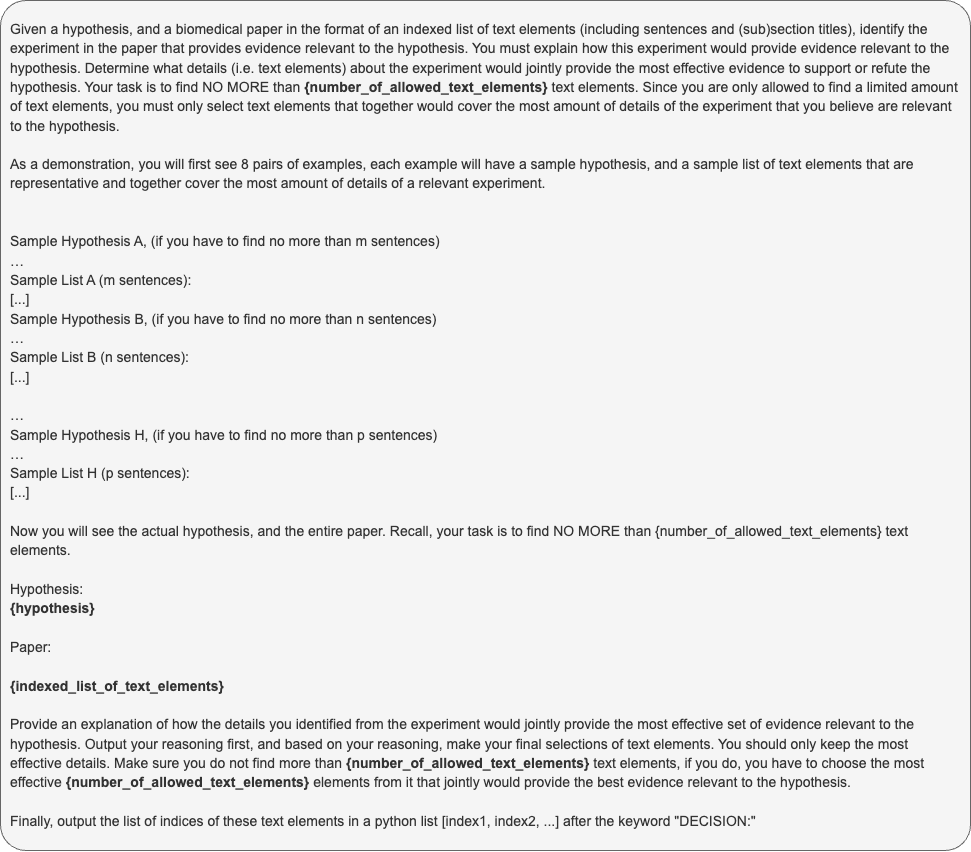}
    \caption{8-shot In-context Learning Prompt template for evaluating LLMs on tasks Evidence Retrieval @Optimal and Evidence Retrieval @10.}
    \label{fig:experiment_icl_default}
\end{figure}

We randomly selected 8 pairs of examples from the development set, which was completely disjointed from the test set. We experimented with different versions of in-context learning. In one attempt, we gave the full paper for each example pair (i.e., sample hypothesis, sample full paper, sample optimal number of or 10 sentences that cover the most amount of study aspects.). However, no LLM improved on the training set using N-shot with full paper, even when N =1 or 2. Therefore, we decided to not use the full paper. Instead, for each example pair, we give only the sample hypothesis and the sample list of sentences that cover the maximum amount of aspects (size = Optimal or 10). See Figure \ref{fig:experiment_icl_default} for its prompt template. 

\newpage
\subsection{Section-by-Section Prompt for Evidence Retrieval @Optimal and @10}
\label{ssec:appendix_secbysec_eroptimal_10}
\begin{figure}[htbp]
    \centering
    \includegraphics[width=\linewidth]{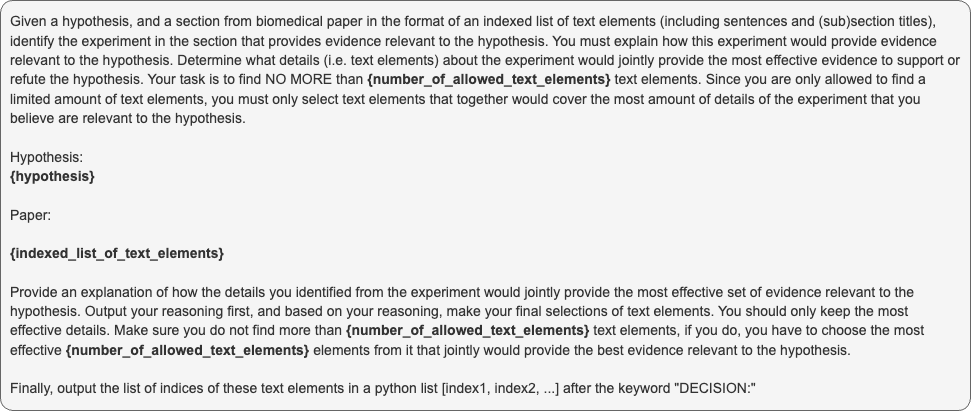}
    \caption{Step 1: Section-by-section Prompt template for evaluating LLMs on tasks Evidence Retrieval @Optimal and Evidence Retrieval @10.}
    \label{fig:experiment_secbysec_default_step1}
\end{figure}

\begin{figure}[htbp]
    \centering
    \includegraphics[width=\linewidth]{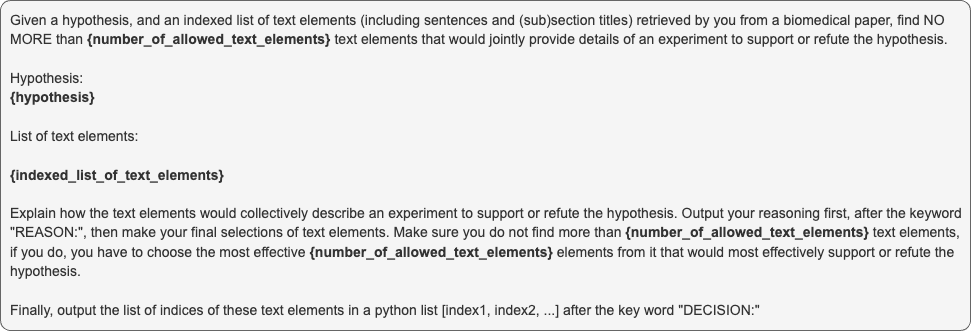}
    \caption{Step 2: Section-by-section Prompt template for evaluating LLMs on tasks Evidence Retrieval @Optimal and Evidence Retrieval @10. The LLM is asked to choose the best K number of sentences.}
    \label{fig:experiment_secbysec_default_step2}
\end{figure}

Section-by-section is a strategy to counter the long-context difficulty posed by EvidenceBench. Instead of processing the full paper at once (typically consisting of more than 5000 tokens), each paper is divided into its naturally defined sections, i.e. introduction, methodology, results, etc. Each time, an LLM only retrieves sentences from one single section. Note, for fairness, for each section, we can only instruct the LLM to retrieve no more than K sentences, even though a paper could have 10 sections. Consequently, the total number of retrieved sentences for all sections combined sometimes exceeds the maximally allowed number of sentences K. Therefore, we have the second step of processing where we ask the model to select the best K sentences from all sentences retrieved from all sections. See Figure \ref{fig:experiment_secbysec_default_step1} for the prompt template of the first step. See Figure \ref{fig:experiment_secbysec_default_step2} for the prompt template that asks the LLM to choose the best K sentences from all its retrieved sentences from all sections. 

\newpage
\subsection{Regeneration Prompt}
\label{ssec:appendix_regeneration prompt}
\begin{figure}[htbp]
    \centering
    \includegraphics[width=\linewidth]{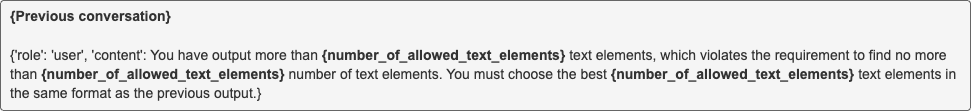}
    \caption{Regeneration prompt template if LLM exceeds the maximally allowed number of text elements.}
    \label{fig:regeneration}
\end{figure}

If an LLM retrieved more than allowed despite explicit instructions, it will be sent a regeneration prompt, with the following format in Figure \ref{fig:regeneration}.

\subsection{Instructions for Embedding Model}
\label{ssec:appendix_instruction_embedding}
For Evidence Retrieval @Optimal and @10, for embedding models, there are two strategies, with instruction or without instruction. The instruction is: 

\texttt{"From a biomedical experiment, find important and representative details that would form the most effective set of evidence relevant to the hypothesis"}

Recall, "with instruction strategy" means concatenating the instruction with the hypothesis and creating an embedding for the concatenated text as a new hypothesis vector. Each text element in the candidate pool (i.e. the paper) is still independently embedded as its own vector.

See Table \ref{tab:embed_models_appendix} for embedding models' performance with instruction and without instruction, on the tasks of Evidence Retrieval @Optimal and @10. 

\begin{table}[ht]
    \footnotesize
        \centering
        \caption{Embedding Models Comparison. Standard Error calculated by bootstrapping.}
        \label{tab:embed_models_appendix}
        \setlength{\tabcolsep}{10pt}
        \begin{tabular}{lcc|cc}
            \toprule
             & \multicolumn{2}{c}{\textbf{No Instruction}} & \multicolumn{2}{c}{\textbf{Instruction}}  \\
            \cmidrule(lr){2-3} \cmidrule(lr){4-5} 
            \textbf{Model} & ER 
            @
            
            optimal & ER @
            
            10 & ER 
            @
            
            optimal & ER @

            10 \\
            \midrule
            BM25 & 16.5 $\pm$ 1.1 & 34.4 $\pm$ 1.7 & - & - \\
            OpenAI & \textbf{25.1 $\pm$ 1.4} & \textbf{42.2 $\pm$ 1.7} & 23.8 $\pm$ 1.4 & 41.3 $\pm$ 1.7 \\
            VoyageAI & 21.6 $\pm$ 1.3 & 42.0 $\pm$ 1.8 & 22.7 $\pm$ 1.3 & 41.9 $\pm$ 1.8 \\
            GritLM & 21.5 $\pm$ 1.2  & 39.7 $\pm$ 1.7& \textbf{27.0 $\pm$ 1.3} & \textbf{46.4 $\pm$ 1.7} \\
            E5-Mistral & 22.7 $\pm$ 1.4 & 41.9 $\pm$ 1.8 & 21.9 $\pm$ 1.3 & 40.8 $\pm$ 1.8\\
            \bottomrule
        \end{tabular}
\end{table}

For tasks such as Results Evidence Retrieval @Optimal and @10, embedding models will take in task-specific instruction. Instruction has the following format:

\texttt{From a biomedical paper, find important and representative details about experiment outcomes, results and analyses that would form the most effective set of evidence relevant to the hypothesis.}

Note, in the instruction, we have concisely and explicitly informed the embedding model that the embedding of the hypothesis should only have high cosine similarity with text elements related to results or analyses. Since we are not changing the embedding for the results and analyses related text elements, we only change the embedding for the hypothesis to fit this purpose. This is an emergent and specially fine-tuned ability of some of the newer and more powerful embedding models, such as GritLM. See Table \ref{tab:comparison_appendix} lower right quadrant for the performance of embedding models with instruction on tasks such as Results Evidence Retrieval (ER) @Optimal and @5.

\newpage
\subsection{Standard Errors for Model Evaluations}
\label{ssec:appendix_standard_error in model evaluation}
In this section, we reproduce the main results from the paper, showing standard errors for all estimates.

In table \ref{tab:comparison_appendix}, we show overall results for model performance on the 4 tasks. For Evidence Retrieval Tasks, three strategies are considered: default, in-context learning, and section-by-section for each LLM. Default refers to Figure \ref{fig:experiment_baseline} prompt. ICL refers to Figure \ref{fig:experiment_icl_default}. Section-by-Section refers to Figure \ref{fig:experiment_secbysec_default_step1} and \ref{fig:experiment_secbysec_default_step2}. For each model, the strategy that achieved the best performance is selected, and that result is reported as the model's performance. For example, for the task ER @optimal, gpt-4o achieves the best aspect recall using ICL, whereas for the task ER @10, gpt-4o achieves the best aspect recall using section-by-section. Note, for embedding models, there are two strategies: with instruction or without instruction. The best performance for each embedding model is recorded.

Note, for the two tasks Results Evidence Retrieval (ER) @Optimal and @5, only one strategy is considered for LLM, which is one-shot ICL with section-by-section processing, see Figure \ref{fig:experiment_baseline_result} and Figure \ref{fig:experiment_baseline_result_step2}.

For Results Evidence Retrieval (ER) tasks, embedding models must only have one strategy, i.e. the strategy with instruction. Since the task is about retrieving sentences related to results and analyses, an embedding model must understand this constraint through its instructions.

\begin{table}[htbp]
    \caption{Aspect Recall for the full retrieval task (top) and the result retrieval task (bottom). For each model, the highest number is reported if multiple strategies are used. Standard Error calculated by bootstrapping. }
    \label{tab:comparison_appendix}
    \centering
    \scriptsize
    \setlength{\tabcolsep}{1pt}
    \begin{tabular}{l|cc|cccc|cccc}
        \toprule
        & Max & Random & GPT-4o & Claude3-Opus & Gemini-1.5 & LLaMA3-70b & OpenAI Emb & VoyageAI & GritLM & E5-Mistral \\
        \midrule
        ER @ optimal & 100.0 & 9.6 & \textbf{51.4 $\pm$ 1.4} & 47.6 $\pm$ 1.5 & 48.3  $\pm$ 1.4 & 46.7 $\pm$ 1.4& 25.1 $\pm$ 1.4& 22.7 $\pm$ 1.3& 27.0 $\pm$ 1.3& 22.7$\pm$ 1.4 \\
        ER @ 10 & 99.3 & 22.3 & \textbf{71.6 $\pm$ 1.5} & 66.4 $\pm$ 1.6& 65.4 $\pm$ 1.6& 65.4 $\pm$ 1.6& 42.2 $\pm$ 1.7& 42.0 $\pm$ 1.8& 46.4 $\pm$ 1.7& 41.9 $\pm$ 1.8 \\
        \midrule
        Result ER @ optimal & 100.0 & 4.4 & \textbf{52.6$\pm$ 2.1} & 51.7$\pm$ 2.1 & 46.7$\pm$ 2.0 & 46.2 $\pm$ 2.2 & 19.1 $\pm$ 1.8& 18.3 $\pm$ 1.8& 18.9 $\pm$ 1.7& 19.3 $\pm$ 1.8\\
        Result ER @ 5 & 99.8 & 11.4 & \textbf{70.8$\pm$ 1.9} & 68.7$\pm$ 2.0 & 65.4$\pm$ 2.0 & 63.7$\pm$ 2.1 & 33.1 $\pm$ 2.2& 31.9 $\pm$ 2.2& 39.1 $\pm$ 2.2& 33.6 $\pm$ 2.2\\
        \bottomrule
    \end{tabular}

\end{table}

\begin{table}[ht]
    \centering
        \centering
        \caption{Comparison of prompting strategies for LLMs. Baseline refers to Figure \ref{fig:experiment_baseline} prompt. ICL refers to Figure \ref{fig:experiment_icl_default}. Sec-by-Sec refers to Figure \ref{fig:experiment_secbysec_default_step1} and \ref{fig:experiment_secbysec_default_step2}. Standard errors are calculated by bootstrapping.
        }
        \label{tab:gen_models_appendix}
         \setlength{\tabcolsep}{3pt}
\begin{tabular}{l>{\centering\arraybackslash}m{1.5cm}>{\centering\arraybackslash}m{1.5cm}|>{\centering\arraybackslash}m{1.5cm}>{\centering\arraybackslash}m{1.5cm}|>{\centering\arraybackslash}m{1.5cm}>{\centering\arraybackslash}m{1.5cm}}
            \toprule
             & \multicolumn{2}{c}{\textbf{Baseline}} & \multicolumn{2}{c}{\textbf{ICL}} & \multicolumn{2}{c}{\textbf{Sec-by-Sec}} \\
            \cmidrule(lr){2-3} \cmidrule(lr){4-5}  \cmidrule(lr){6-7}
            \textbf{Model} & ER 
            @
            
            optimal & ER@
            
            10 & ER 
            @
            
            optimal & ER@
            
            10 & ER 
            @
            
            optimal & ER@
            
            10 \\
            \midrule
            GPT-4o & \textbf{48.1 $\pm$ 1.5} & \textbf{69.6 $\pm$ 1.5} & \textbf{51.4 $\pm$ 1.4} & \textbf{68.7 $\pm$ 1.6} & \textbf{50.9 $\pm$ 1.4} & \textbf{71.6 $\pm$ 1.5} \\
            Claude3-opus & 41.1 $\pm$ 1.6 & 53.6 $\pm$ 1.6 & 38.3 $\pm$ 1.4& 55.4 $\pm$ 1.7 & 47.6 $\pm$ 1.5 & 66.4 $\pm$ 1.6 \\
            Gemini-1.5 & 42.7 $\pm$ 1.5 & 63.0 $\pm$ 1.6 & 43.2 $\pm$ 1.5 & 62.4 $\pm$ 1.7 & 48.3 $\pm$ 1.4 & 65.4 $\pm$ 1.6 \\
            LLaMA3-70b & - & - & - & - & 46.7 $\pm$ 1.4 & 65.4 $\pm$ 1.6 \\
            \bottomrule
        \end{tabular}
\end{table}

Table \ref{tab:gen_models_appendix} shows model performance for the three prompting strategies.

We observe that all standard errors are less than 2\%, indicating we have a sufficiently large test set size to effectively distinguish different models and prompting strategies.

\newpage
\subsection{Negation Prompt}
\label{ssec:appendix_negation prompt}
\begin{figure}[htbp]
    \centering
    \includegraphics[width=\linewidth]{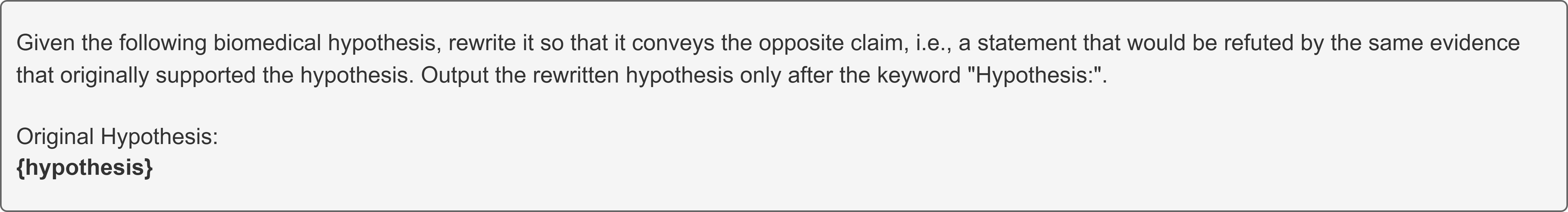}
    \caption{The prompt to negate a hypothesis}
    \label{fig:negation prompt}
\end{figure}

We evaluate models on negated hypotheses using prompt in Figure \ref{fig:negation prompt}. We give two example pairs of original hypothesis and negated hypothesis:

\begin{itemize}
    \item \textbf{Original Hypothesis:} Glucocorticoids cause cleft palate by inhibiting growth of the palatal shelves, preventing shelf contact and fusion.

    \textbf{Negated Hypothesis:} Glucocorticoids prevent cleft palate by promoting growth of the palatal shelves, facilitating shelf contact and fusion.

    \item \textbf{Original Hypothesis:} The mortality rate from bladder cancer is higher among painters than in the general population.

    \textbf{Negated Hypothesis:} The mortality rate from bladder cancer is lower among painters than in the general population.
\end{itemize}

\section{Model Sensitivity to Paraphrased Hypothesis}
\label{sec:appendix_model sensitivity to hypothesis}
We test models' sensitivity to different paraphrased versions of the hypothesis. We use a variety of LLMs to paraphrase a hypothesis. Note, a paraphrase would still keep the scientific terminology to make sure it is still the same hypothesis. As shown in Figure \ref{tab:paraphrased_hypothesis_comparison}, GPT-4o and the two embedding models are less sensitive to paraphrased hypotheses, while Claude3-Opus is more sensitive to them.

\begin{table}[htbp]
    \caption{\small{Models evaluated under paraphrased versions of the hypothesis. All experiments are under task Results ER @ optimal.}}
    \label{tab:paraphrased_hypothesis_comparison}
    \centering
    \setlength{\tabcolsep}{4pt}
    \begin{tabular}{l|c|c|c|c|c}
        \toprule
        Model & Original & GPT-4o & Claude & Llama3-70B & Llama3-8B \\
         & Hypothesis & Paraphrased & Paraphrased & Paraphrased & Paraphrased\\
        \midrule
        GPT-4o            & 52.6 & 49.9 & 51.7 & 50.8 & 50.3 \\
        Claude3-Opus      & 51.7 & 44.6 & 41.9 & 43.3 & 43.9 \\
        GritLM            & 18.9 & 22.8 & 19.4 & 20.8 & 19.1 \\
        OpenAI Emb        & 19.1 & 19.1 & 18.1 & 19.7 & 18.3 \\
        \bottomrule
    \end{tabular}
\end{table}

\newpage
\section{Hypothesis Collection}
\label{sec:appendix_hypothesis generation}
\begin{figure}[htbp]
    \centering
    \includegraphics[width=\linewidth]{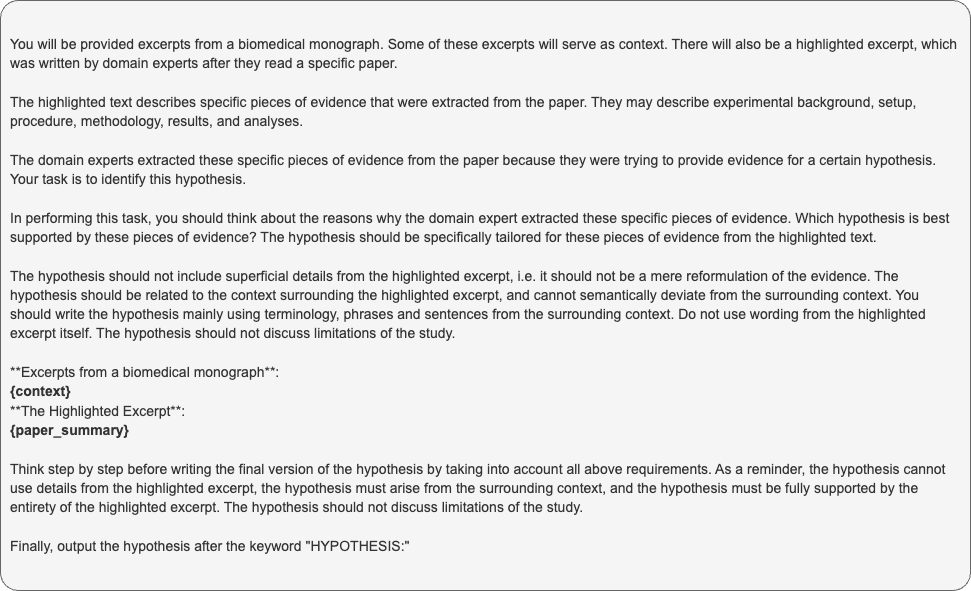}
    \caption{First step of hypothesis collection from review papers}
    \label{fig:hypothesis_generation_step1}
\end{figure}

There are two steps for collecting hypotheses from the review papers. \\
\textbf{Step 1}:\\
See Figure \ref{fig:hypothesis_generation_step1} for the prompt template. Note that Claude3-Opus is used for this step. The model is instructed to answer, "What is the motivation for the expert reviewer to extract these pieces of information from the paper and summarize them?". 
\vspace{-10pt}
\vspace{-10pt}

\begin{figure}[htbp]
    \centering
    \includegraphics[width=\linewidth]{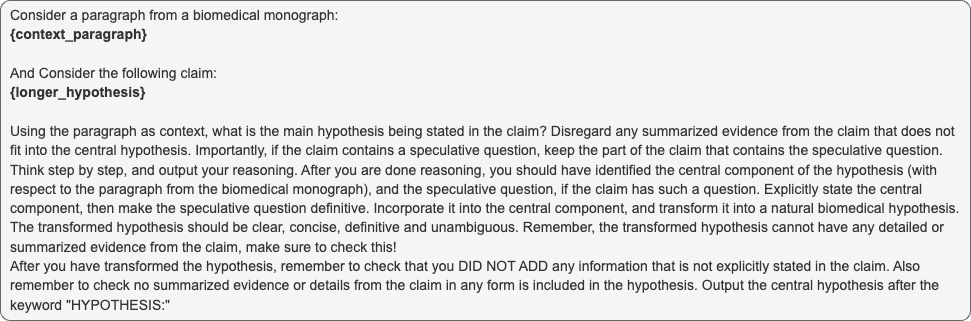}
    \caption{Second step of hypothesis collection, trimming the output of step 1.}
    \label{fig:hypothesis_generation_step2}
\end{figure}

\textbf{Step 2:}\\
In order to make sure the hypotheses have the correct format and do not have any summarized evidence, we perform a second step where we trim the first-step output off any unnecessary details and specifications and only preserve the central biomedical question. See Figure \ref{fig:hypothesis_generation_step2} for the prompt template.

\vspace{-10pt}
\newpage
\section{Data Preprocessing}
\label{sec:appendix data preprocessing}
\subsection{Harvesting Evidence Summary}
\label{ssec:appendix harvesting evidence summary}
We extract expert-written evidence summaries from review papers using an algorithmic procedure. Summaries that cite multiple papers are excluded. We also exclude any paper that is not licensed under Creative Commons or is not in the public domain. Finally, we remove any paper that is cited twice in one section of a review paper, ensuring that the extracted evidence summaries are complete.

\subsection{Further Processing Evidence Summary}
\label{ssec:appendix_extracting evidence summary}
\begin{figure}[htbp]
    \centering
    \includegraphics[width=\linewidth]{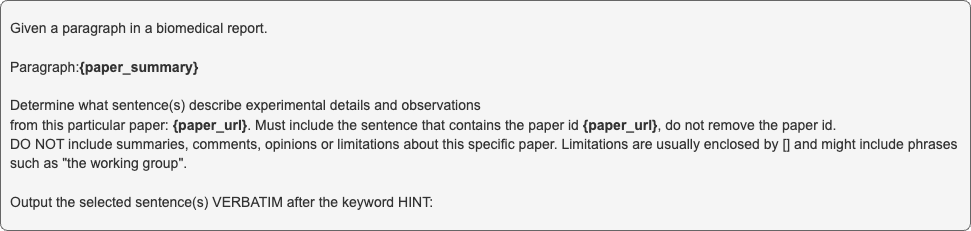}
    \caption{Prompte for extracting evidence summary for a cited paper}
    \label{fig:extract_summary}
\end{figure}

 The primary difficulty in ensuring the high quality of evidence summary extracted by the algorithm lies in the variability in the placement of the citation (e.g., \cite{aristolochic_acid}). Since review papers sometimes have hundreds of papers cited in one section, each with a evidence summary surrounding the citation, it is challenging to heuristically delineate the boundaries of the evidence summaries, and motivates the use of an LLM for this step.
See Figure \ref{fig:extract_summary} for prompt template

\subsection{Identifying Suitable evidence summaries}
\label{ssec:appendix_identifying suitable evidence summary}
\begin{figure}[htbp]
    \centering
    \includegraphics[width=\linewidth]{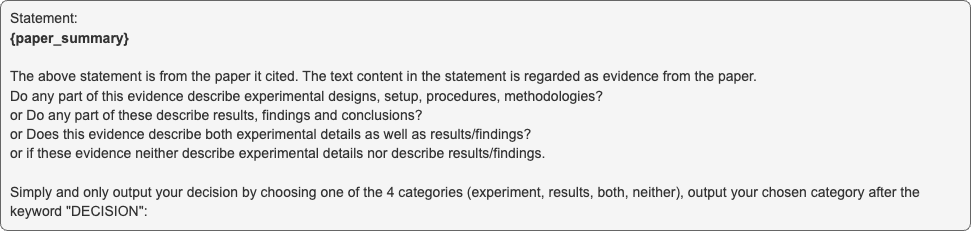}
    \caption{Prompt template to determine if a evidence summary is suitable for EvidenceBench. Here, "both" or "results" is acceptable.}
    \label{fig:specificity_of_paper_summary}
\end{figure}

An evidence summary that is suitable for EvidenceBench contains experimental outcomes, results, analyses, or methodology. However, some evidence summaries only have high-level information about a paper and do not have the desired level of specificity. We use an LLM to determine if a evidence summary suits EvidenceBench. See Figure \ref{fig:specificity_of_paper_summary} for prompt template. 

\subsection{Human Verification}
\label{ssec:human verification of harvest}
We randomly sampled 50 extracted evidence summary using the entire harvesting procedure and LLM preprocessing. Only in 1 case did we notice where the extracted evidence summary includes a sentence that does not belong to this specific paper. This ensures the high quality and accuracy of the harvesting procedure and LLM preprocessing.

\section{Fine-tuning and Evaluation on EvidenceBench-100k}
\label{appendix:finetuning}
EvidenceBench-100k also split into a 80k train set and a 20k test set. For cost reasons, we random sample 300 datapoints from the 20k test set. We evaluated various representative LLMs and embedding models on the Task of Result ER@Optimal on this new 300 points test set from EvidenceBench-100k. From table \ref{table:new_test_set}, we clearly see LLMs dominate over embedding models, while GritLM-7B outperforms other embedding models. This shows EvidenceBench-100k test set can also be used to evaluate and differentiate various models' performance.

Furthermore, to demonstrate the quality of EvidenceBench-100k's training set, we fine-tuned two models. E5-v2 335M is a light-weight embedding model. Since there are over 100 million sentence-aspect pair judgments from the train set,  we randomly sampled 1 million triplets, where each triplet contains an anchor (the study aspect), a positive (a sentence that is considered as source of information for the study aspect), a negative (a sentence that is not considered as source of information for the aspect). Using a margin-based triplet-loss with margin=0.05, AdamW optimizer  (weight decay =0.01), peak learning rate of 5e-4, 10\% linear warmup then cosine-annealing, batch size =256, we trained E5-v2 with full-parameter tuning for one epoch. We also fine-tuned Llama3-8B using all 80k training datapoints. Llama3-8B is trained with LoRA rank=8, alpha =16, a batch size of 16, AdamW optimizer (weight decay =0.01), and the same learning rate scheduler, we fine-tuned the model for one epoch. Note, during inference time we trained LLama3-8B to only output sentence indices, during training, we trained it to output both sentence indices, sentence texts and the summarized list of study aspects. This has shown to be more effective than only training it on sentence indices. 

We made sure none of the papers in the original EvidenceBench is used in any way in the EvidenceBench-100k to avoid contamination. Our fine-tuned E5 and Llama3-8B are tested on the original EvidenceBench test set for the task of Result ER@Optimal. 

Notice that the fine-tuned E5 model has dramatically improved its performance on the original test set, surpassing all larger embedding models, see Table \ref{table:fine_tuned}. It only trails behind the performance of large language models. The fine-tuned Llama3-8B has a performance of 41.0\% which trails behind much larger SoTA LLMs. This shows that EvidenceBench-100k is suitable for developing and training both embedding-based information retrieval systems as well as large language models. 

\section{Qualitative Analysis for GPT-4o on the Original EvidenceBench}
\label{appendix:qualitative analysis for gpt-4o}

For the task Result ER@ Optimal, there are a total of 1228 results study aspects that the review papers identified. 
 
Out of these 1228, GPT-4o's retrieved sentences fail to cover 583 study aspects, from which 50 pairs (missed "Result" aspect, set of GPT-4o retrieved sentences) are randomly sampled and manually inspected by two researchers. 
Out of the 50 cases, 10 cases should not be considered GPT-4o errors. In 7 cases, GPT-4o retrieved an almost sufficient set of sentences but missed one aspect due to the upper limit on the number of sentences. This issue arose from a failure in the optimization step during retrieval, not from a reasoning or retrieval error. In 1 case, our alignment annotation procedure missed one of the GPT-4o retrieved sentences as the source of information for a study aspect. In 1 case, a parsing error occurred where our algorithm did not extract the full summary from the review paper. Finally, in 1 case, one method-related study aspect is mistakenly labeled as a result-related study aspect.

\newpage
\section{Further Analyses}
\label{sec:analysis}

\textbf{Effectiveness of Current LLM Solutions:}
Table \ref{tab:comparison} clearly indicates that current embedding models are inadequate for assisting or replacing human experts in identifying relevant evidence for biomedical hypotheses. We now examine the best-performing LLM solution, GPT-4o. In the task ER@Optimal, GPT-4o retrieves an average of 4-5 sentences per hypothesis, according to Table \ref{tab:testset statistics}. In this setting, GPT-4o achieves a 50\% Aspect Recall, covering half of the study aspects identified by human experts. This demonstrates that current LLMs cannot fully replace human experts in finding relevant evidence for hypotheses.

Conversely, per the task definition for ER@10, models retrieve 10 sentences. In this setting, GPT-4o achieves an aspect recall of 70\%. According to Table \ref{tab:testset statistics}, a typical research paper contains 168 sentences. Therefore, instead of reviewing the entire research paper, humans can use the 10 sentences retrieved by GPT-4o as an efficient starting point, while searching for additional sentences that cover potentially missing aspects. This demonstrates that, in this setting, GPT-4o can meaningfully assist human experts in locating and presenting evidence from research papers for hypotheses, a crucial step in writing review papers.

\textbf{Embedding Models Underperform Generative Models:}
From Table \ref{tab:comparison}, we observe that the performance of embedding models falls significantly short of the performance of similar-sized generative models. The embedding models GritLM-7B, E5-Mistral-7B and the state-of-the-art NV-Embed-7B cover at most 20.1\% of aspects, while the pretrained LLama3-8B covers 35.8\% of aspects. 
The shortcomings of GritLM-7B and NV-Embed-7B suggest that a naive local embedding of sentences, without contextual awareness, is insufficient for this task. Our empirical observations confirm that reasoning beyond individual sentences is necessary to solve this task effectively. For instance, in Figure \ref{fig:evaluation}, Sentence 9 and Sentence 69 convey the same information and both address study aspect 1. Only by comparing them together (i.e., reasoning beyond a single sentence) can models eliminate one of these sentences to reduce redundancy. 

\textbf{Section-level Reasoning is Sufficient:}
Table \ref{tab:gen_models} shows that retrieving evidence section-by-section (Sec-by-Sec) achieves strong performance on the task. With Sec-by-Sec, a model can only read one section at a time, preventing it from reasoning across multiple sections. The strong performance of this method indicates that global reasoning across the entire research paper is not essential for retrieving evidence. This can be explained by the structure and organization of biomedical research papers, where the content of each section is relatively self-contained, and interaction across sections is sparse.

This suggests that LLMs trained for much longer contexts (e.g., over 10,000 tokens) may not be necessary for this task.

\begin{figure}[t]
    \centering
    \includegraphics[width=\linewidth]{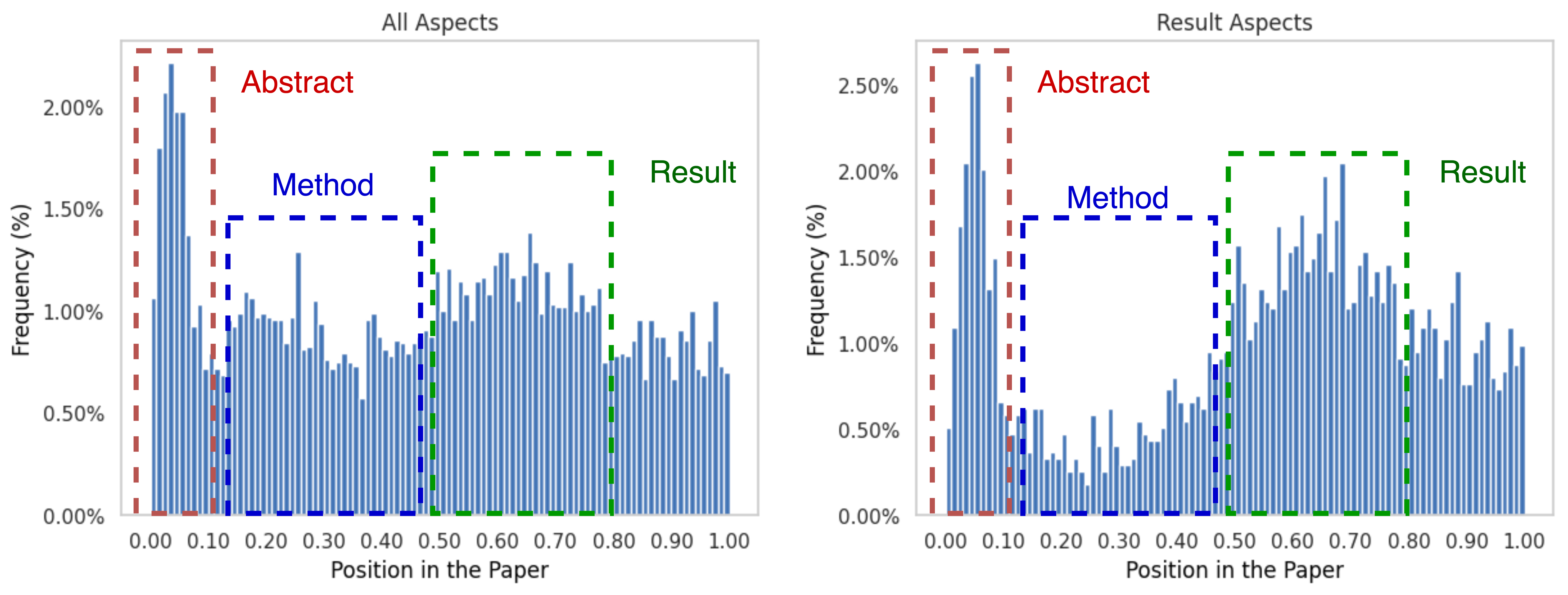}
    \caption{\small{Original EvidenceBench. \textbf{Left} figure shows the distribution for the relative position in the candidate pool of all sentences that are considered as source of information for at least one study aspect. \textbf{Right} figure shows the same but for aspects labeled as 'Results'. Abstract sentences have much higher chance to be matched with aspects. However, all abstract sentences (around 10 sentences per research paper) only cover around 50\% of all aspects, indicating no heuristic algorithm can cheat EvidenceBench.}}
    \label{fig:distribution}
\end{figure}

\textbf{LLMs Still Get "Lost in the Middle".} We observed that some papers have most of their important sentences concentrated at the beginning and end of the document, as shown in Figure \ref{fig:distribution}. We categorize these papers into two groups. In the first category, the first 10 and last 10 sentences of a paper cover over 80\% of the study's aspects. In the second category, these sentences cover less than 20\% of the aspects. Out of 3,000 randomly sampled points from the EvidenceBench-100k test set, 1,115 papers fall into the first category, while 1,111 papers fall into the second category. GPT-4o's aspect recall performance is 51.6\% in the first category and 34.9\% in the second category. Claude3-Opus aspect recall is 42.8\% in the first category and 28.2\% in the second category. Note, during evaluation, we neither explicitly nor implicitly instruct LLMs to focus on any specific parts of the paper. 

This indicates that LLMs perform significantly better when the important tokens are not located in the middle of the document. Our manual inspection also reveals that LLMs tend to focus on the beginning and end of the document. The "Lost in the Middle" phenomenon, as reported in previous LLMs \citep{lost_in_the_middle}, seems to persist in current LLMs on EvidenceBench.

\clearpage

\end{document}